%% file: arxiv.tex
\documentclass[10pt,twocolumn,letterpaper]{article}

\usepackage{cvpr}      


\definecolor{cvprblue}{rgb}{0.21,0.49,0.74}
\usepackage[pagebackref,breaklinks,colorlinks,allcolors=cvprblue]{hyperref}
\usepackage{multirow}
\usepackage{pifont}
\usepackage{xcolor}
\usepackage{tcolorbox} 
\usepackage{booktabs}
\usepackage{graphicx}
\usepackage{times}
\usepackage{epsfig}
\usepackage{graphicx}
\usepackage{amsmath,amsfonts,amssymb}
\usepackage[ruled,vlined]{algorithm2e} 
\usepackage{caption}
\usepackage{xcolor} 
\usepackage{hyperref} 
\hypersetup{
    colorlinks=true,
    linkcolor=red,
    filecolor=red,
    linkbordercolor=red,
}
\usepackage{tablefootnote}

\title{DrVideo: Document Retrieval Based Long Video Understanding}

\author{
 Ziyu Ma*\textsuperscript{1,2},
 Chenhui Gou*\textsuperscript{2},
 Hengcan Shi\textsuperscript{1,2},
 Bin Sun\textsuperscript{1},
 Shutao Li\textsuperscript{1},\\
 Hamid Rezatofighi\textsuperscript{2},
 Jianfei Cai\textsuperscript{2}
\\
 \textsuperscript{1} College of Electrical and
Information Engineering, Hunan University,\\
 \textsuperscript{2} Data Science \& AI Department, Faculty of IT, Monash University
\\
 \texttt{
 {\{maziyu, sunbin611, shutao\_li\}@hnu.edu.cn}}\\
 \texttt{
 {\{chenhui.gou, hengcan.shi, hamid.rezatofighi, jianfei.cai\}@monash.edu}
 }\\
\small{* Equal contribution }\\
\vspace{-1em}
}
\begin{document}
\maketitle
\input{sec/0_abstract}
\input{sec/1_intro}

\input{sec/2_related_work}
\input{sec/3_methods}

\input{sec/4_experiment}

\clearpage
\clearpage
\input{sec/X_suppl}
\clearpage
\begingroup
\small
\bibliographystyle{ieeenat_fullname}
\bibliography{arxiv}
\endgroup
\end{document}

%% file: sec/0_abstract.tex
\begin{abstract}
Most of the existing methods for video understanding primarily focus on videos only lasting tens of seconds, with limited exploration of techniques for handling long videos. The increased number of frames in long videos poses two main challenges: difficulty in locating key information and performing long-range reasoning. Thus, we propose DrVideo, a document-retrieval-based system designed for long video understanding. Our key idea is to convert the long-video understanding problem into a long-document understanding task so as to effectively leverage the power of large language models. Specifically, DrVideo first transforms a long video into a coarse text-based long document to initially retrieve key frames and then updates the documents with the augmented key frame information. 
It then employs an agent-based iterative loop to continuously search for missing information and augment the document until sufficient question-related information is gathered for making the final predictions in a chain-of-thought manner. Extensive experiments on long video benchmarks confirm the effectiveness of our method. DrVideo significantly outperforms existing LLM-based state-of-the-art methods on EgoSchema benchmark (3 minutes), MovieChat-1K benchmark (10 minutes), and the long split of Video-MME benchmark (average of 44 minutes).
\end{abstract}

%% file: sec/1_intro.tex
\section{Introduction}

Video understanding is a challenging task in computer vision, requiring the processing of spatio-temporal information and advanced reasoning abilities. Previous works have successfully processed short videos lasting around tens of seconds~\cite{yang2021just, VindLU_CVPR2023, lei2021less, yang2022zero, suris2023vipergpt, lin2023mm}. 
However, how to deal with long video understanding remains unclear. Recent advancements in large language models (LLMs) ~\cite{instructGPT, chatgpt, gpt3, bloom, llama, chowdhery2022palm, ma2024gerea} demonstrate strong abilities in language understanding and reasoning across long text sequences. These advancements have inspired the development of video-language models (Video-LMs) \cite{videoChatGpt, zhang2024llavanextvideo, kim2024image, xu2024pllava} to address long video understanding issues. 

Video-LMs typically encode a video as a sequence of visual tokens, which are then concatenated with language tokens into a long sequence and LLM is then used to understand this long sequence. Although Video-LMs improve video understanding, they have the following limitations. (i) They cannot process an entire video as input to a Video-LM (\textit{e.g.,} the visual encoder OpenAI’s CLIP-L-14 outputs 24×24 tokens for each image, while the max length of LLaVA-NeXT-Video \cite{zhang2024llavanextvideo} is only 8192). (ii) They typically perform uniform or random frame sampling at large strides to handle long videos without considering the content (\textit{e.g.,} 16 for PLLaVA \cite{xu2024pllava} and 6 for IG-VLM \cite{kim2024image}), leading to potential key information loss. (iii) The simple concatenation of visual tokens increases the difficulty for the LLM to locate question-related (key) visual information within the long video sequence, complicating long-range reasoning across the vision token sequences.\par
\begin{figure*}[t]
    \centering
    \includegraphics[scale=0.63]{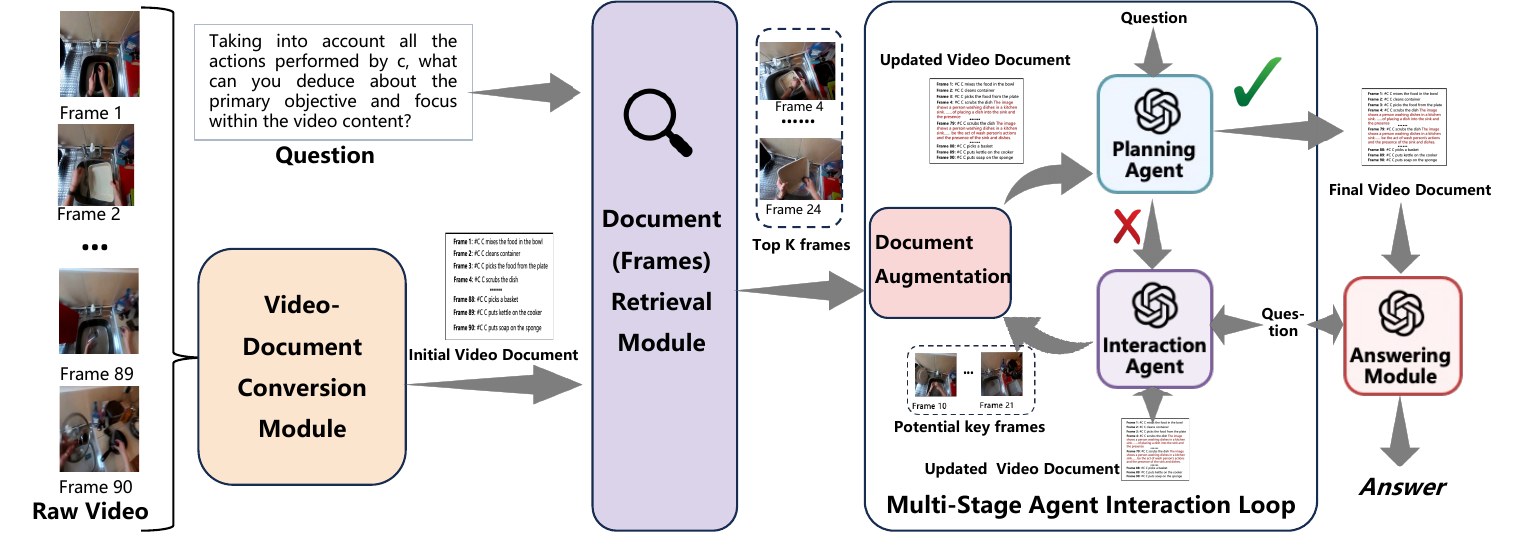}

    \caption{\textbf{Overview of our DrVideo framework.} It comprises five components: a video-document conversion module, a retrieval module, a document augmentation module, a multi-stage agent interaction loop and an answering module.}
    \label{fig1}
\end{figure*}
Apart from the studies of encoding a video as a sequence of visual tokens, another line of research converts raw videos into captions and then utilities the long-range reasoning abilities of LLMs to predict the final answer. A pioneering work - LLoVi \cite{LLoVi} uses a short-term video model \cite{zhao2023learning} in conjunction with an LLM \cite{gpt3.5} to solve the 
long video understanding task. Given a long video, LLoVi first divides it into multiple short clips and converts them into short textual descriptions. Afterward, these short textual descriptions are summarized by the LLM, and the summary is finally used by LLoVi to answer the given question. 
Inspired by LLoVi, VideoAgent \cite{wang2024videoagent} designs an agent-based system to further leverage the reasoning abilities of LLMs and locate question-related video clips or images. Initially, VideoAgent samples only a minimal number of frames from the video (\textit{i.e.,} 5 frames), converting them into textual descriptions as input for the LLM. Then, through continuous feedback and interaction with the LLM \cite{openai2023gpt4}, VLM \cite{zhao2023learning}, and CLIP \cite{sun2024eva}, it gradually finds question-related frames in a coarse-to-fine manner via image-text similarity. Finally, VideoAgent answers the question based on these question-related frames. Despite these LLM-based methods overcoming the challenge of performing long-range reasoning and achieving promising results, they still suffer from the following problems.

\begin{itemize}
\item These methods struggle to accurately identify question-related frames. For instance, VideoAgent \cite{wang2024videoagent} leverages the LLM to infer missing question-related information based on prior knowledge and current key frame information, then uses CLIP’s image-text similarity to select the most likely frames. However, such a coarse-to-fine method lacks a holistic grasp of video content and overly depends on the LLM’s priors, often failing in locating question-relate frames, particularly in long videos.
\item These methods convert videos into captions, addressing long-range reasoning but overlooking critical information loss. Specifically, even if question-related frames are identified, the generated captions cannot cover all the essential information in the key frames. For example, the caption \textit{``a woman is looking in the mirror''} does not contribute to answering the question \textit{``what is the woman wearing when standing in front of the mirror in the video''}. In this case, LLM can only make an aimless and biased surmise to answer the question.
\end{itemize}

To solve the above problem, we turn to how human understands a long video. In particular, a human needs to gain a general understanding of the video, identifying events across segments. Then, the human locates question-related segments, examines them in detail, and answers the question based on the extracted information. Inspired by this, 
in this paper, we propose DrVideo, a document retrieval-based system for long video understanding that progressively finds key frames and generates a question-related augmented document for LLMs to make confident predictions. Our key idea is to convert long video understanding into long document understanding and leverage the impressive capabilities of LLMs in key information location \cite{perot2023lmdx} and long-range reasoning \cite{ding2024longrope} on various NLP tasks.

Specifically, DrVideo transforms raw long videos into long documents using a pretrained frame/clip-level visual captioner (\textit{e.g.,} LLaVA-NeXT \cite{liu2024llavanext}) to make the LLM have a holistic grasp of video content. Different from VideoAgent \cite{wang2024videoagent}, which identifies key frames via image-text similarity without information augmentation, DrVideo proposes a retrieval module and a multi-stage agent interaction loop that dynamically finds the potential missing information through sequential steps of reasoning and then interacts with a document augmentation module to request augmentation of the information within language space. Once sufficient question-related information is gathered, it feeds into the answering module to get the final prediction. \par

DrVideo achieves significant performance gains (3.8\% on EgoSchema subset \cite{mangalam2023egoschema}, 8.3\% on MovieChat-1K global mode, 24.8\% on MovieChat-1K breakpoint mode \cite{song2023moviechat}, and 6.3\% on the long split of Video-MME \cite{fu2024video}) over the existing LLM-based state-of-the-art methods \cite{LLoVi,fan2024videoagent,wang2024videoagent,min2024morevqa} across different long video benchmarks, from 3 minutes to 10 minutes and longer than 1 hour. Surprisingly, combining the subtitles on the long split of Video-MME benchmark, DrVideo obtains 71.7\% accuracy, outperforming many heavily-engineered large-scale proprietary models (\textit{e.g.,} Gemini 1.5 Flash \cite{team2024gemini}, GPT-4o mini \cite{openai2024gpt4o}, and GPT-4V \cite{openai2023gpt4v}). This highlights the great potential of DrVideo. Moreover, DrVideo is a training-free framework and is researcher-friendly, as our results can be replicated on an RTX 4090 with a reasonable number of GPT accesses.\par
The contribution of this work can be summarized as:
\begin{itemize}
\item We propose DrVideo, the first document-retrieval-based system designed for the long video understanding task by converting it into a long-document retrieval and understanding task to effectively leverage the power of large language models.
\item Different from previous works, we propose a retrieval module and a new multi-stage agent interaction loop that dynamically finds the potential missing information and augments these information within language space.
\item DrVideo outperforms the existing LLM-based state-of-the-art methods across different long video benchmarks (\textit{i.e.,} EgoSchema \cite{mangalam2023egoschema}, MovieChat-1K \cite{song2023moviechat}, and the long split of Video-MME \cite{fu2024video}) by a margin.
\end{itemize}

%% file: sec/2_related_work.tex
\section{Related Work}
\noindent \textbf{Long Video Understanding.} Modeling long videos, which are several minutes or more in length, generally requires advanced temporal modeling, resulting in complex model designs. LF-VILA~\cite{sun2022long} introduces a Temporal Window Attention (HTWA) mechanism to capture long-range dependencies in long videos. MeMViT~\cite{wu2022memvit} and MovieChat~\cite{song2023moviechat} employ a memory-based design to save question-related information from previous video segments. Some other approaches employ space-time graphs~\cite{hussein2019videograph,wang2021supervoxel} or relational space-time modules~\cite{yang2023relational} to capture spatio-temporal dependencies from the raw long videos. Recently, S4ND~\cite{nguyen2022s4nd}, ViS4mer~\cite{islam2022long}, and S5~\cite{wang2023selective} have used Structured State-Space Sequence (S4)~\cite{gu2021efficiently} layers to capture long-range dependencies in videos. Different from these methods, our DrVideo does not design a complex module to perform long video understanding. Instead, we develop a document retrieval-based system with LLM for zero-shot long video understanding.

\noindent \textbf{LLMs for Video Understanding.} 
The recent rise of LLMs \cite{zeng2022socraticmodels} and VideoChat~\cite{li2023videochat} align the visual features extracted by pretrained visual models to LLMs and apply them to video understanding. Video ChatCaptioner~\cite{chen2023video} and ChatVideo~\cite{wang2023chatvideo} utilize LLMs to represent videos and engage users through dialogues, respectively. VidIL~\cite{wang2022language} applies the image-level models to video understanding tasks via few-shot learning. In addition to short-term video understanding, recent studies~\cite{lin2023mm, chung2023long,bhattacharya2023video} have explored LLMs for long-range video modeling. For instance, GPT-4 is applied in various long-range video modeling tasks in~\cite{lin2023mm}, though quantitative evaluation is limited. Meanwhile, the research in~\cite{chung2023long} focuses on movie datasets with minimal visual analysis~\cite{mangalam2023egoschema}, relying largely on speech and subtitles. In contrast, DrVideo focuses on vision modality in multiple benchmarks.

\noindent \textbf{LLM Agents.} 
In parallel, the computer vision community has started exploring the use of LLMs as agents in various vision tasks such as GUI understanding and robot navigation~\cite{suris2023vipergpt,hong2023cogagent,driess2023palm,brohan2023rt}. In the realm of long video comprehension, initial efforts have employed an agent-like approach, where LLMs interact with external tools or integrate additional functionalities~\cite{suris2023vipergpt,gao2023assistgpt,yang2024doraemongpt,wang2024videoagent,fan2024videoagent}. In contrast to these approaches, our method, DrVideo, reconceptualizes long video understanding as a process of document retrieval, augmentation, and understanding, to leverage the strong capability of LLMs. 


%% file: sec/3_methods.tex
\section{Methodology}

In this section, we detail our proposed document retrieval-based system for long video understanding. As illustrated in Figure \ref{fig1}, given a long video and a question about the video, DrVideo first translates the long video into a long document, referred to as the initial video document. Then, the retrieval module identifies the top $K$ key frames by calculating the similarity between the question and the initial video document. The document augmentation module enriches the information of these key frames and adds it to the initial video document, creating a new video document (named the updated video document), which is used as the starting point for the subsequent multi-stage agent interaction loop.

The loop contains two distinct agents: the planning agent and the interaction agent. The planning agent judges whether the updated video document is sufficient for answering the question. If not, the updated video document is fed into the interaction agent to dynamically find missing key frames. The interaction agent interacts with the document augmentation module to request the required information of these key frames. The output of the document augmentation module is added to the current document to obtain the latest updated video document. The new video document is then sent to the planning agent again to further judge whether the current information is sufficient. This searching and interaction loop continues until the planning agent considers the current information sufficient for answering the question or the maximum iteration is reached. After the loop ends, the final video document and the related question are given to the answering module to get the final prediction. 



\subsection{Video-Document Conversion Module }
By translating each single video frame $V_t$ (or a short video clip) into a short description 
$S_{V_t}$, we convert an input long video into the initial video document ${Doc}_{init}$: 
 \begin{equation}
\resizebox{0.8\linewidth}{!}{${Doc}_{init} = \left\{ \{1, S_{V_1}\}, \{2, S_{V_2}\}, \ldots, \{T, S_{V_T}\} \right\}$}
\label{equ:doc1}
\end{equation}
where $T$ is the length of the video. 
Specifically, $\text{S}_{V_t}$ is generated by a large vision-language captioning model  $\phi_{vlm}$ (\textit{e.g.,} LLaVA-NeXT \cite{liu2024llavanext})
, \ie, $\text{S}_{V_t} = \phi_{vlm}(\mathcal{P},V_t)$, where $\mathcal{P}$ is the prompt for requesting a short description about the image, \eg, \textit{describe the picture in no more than 50 words}. The captioning model used in DrVideo can be replaced by other captioning models (\textit{e.g.,} LaViLa \cite{zhao2023learning}, BLIP-2 \cite{li2023blip2}). 
The experimental results with different captioning models can be found in Table~\ref{tab6}.

\subsection{Document (Frames) Retrieval Module }
After obtaining 
$Doc_{init}$, we introduce a document retrieval module to identify the top question-related frames by calculating the similarity between the question and the whole document. Specifically, we use the OpenAI embedding model $\phi_{emb}$ \cite{openai2024} to obtain the vector representation of the initial document, \ie, $\mathcal{E}_\text{doc} = \phi_{emb}(Doc_{init})$. 
Given a specific retrieval text $\mathcal{RT}$, \ie, the question $Q$, the retrieval module first computes the embedding \(\mathcal{E}_{\mathcal{RT}} = \phi_{emb}(\mathcal{RT)} \), 
and then retrieves the top $K$ frames based on the cosine similarity between $\mathcal{E}_{\mathcal{RT}}$ and $\mathcal{E}_\text{doc}$:
\vspace{-0.2cm}
\begin{equation}
\resizebox{0.7\linewidth}{!}{${topk\_doc} = \underset{t}{\text{arg\,top$K$}}\cos(\mathcal{E}_{\mathcal{RT}}, \mathcal{E}_{\text{doc}_t})$}
\vspace{-0.2cm}
\end{equation}
where $\mathcal{E}_{\text{doc}_t}$ is the embedding of the $t$-th description $S_{V_t}$. 




\subsection{Document Augmentation Module} 
For each $t' \in {topk\_doc}$, we use the LLaVA-NeXT model~\cite{liu2024llavanext} with different prompts $\mathcal{AP}$ (augmented prompts) to generate a detailed description $L_{V_{t'}}$, \ie, $L_{V_{t'}} = \phi_{vlm}(\mathcal{AP}_{t'}, V_{t'})$. 
The initial $\mathcal{AP}_{t'}$ is a general prompt: \textit{If there are factual errors in the question, provide a precise description of the image; if not, proceed to answer the question: $\{Q\}$}. The updated video document $\mathcal{AD}$ becomes
\begin{equation}
\resizebox{0.8\linewidth}{!}{$
\mathcal{AD} = \left\{ 
{\begin{cases} 
\{t, S_{V_t}, L_{V_t'}\} & \text{if } t = t' \\
\{t, S_{V_t}\} & \text{otherwise}
\end{cases} }\bigg| t = 1, \ldots, T \right\}$}.
\end{equation}
Note that $K$ is significantly smaller than the total number of frames (\eg, $K$ = 5 compared to $T$ = 90), so the additional descriptions will not substantially increase the overall length of the document. 

\subsection{Multi-Stage Agent Interaction Loop}
\begin{figure}[t]
    \centering
    \includegraphics[width=\linewidth]{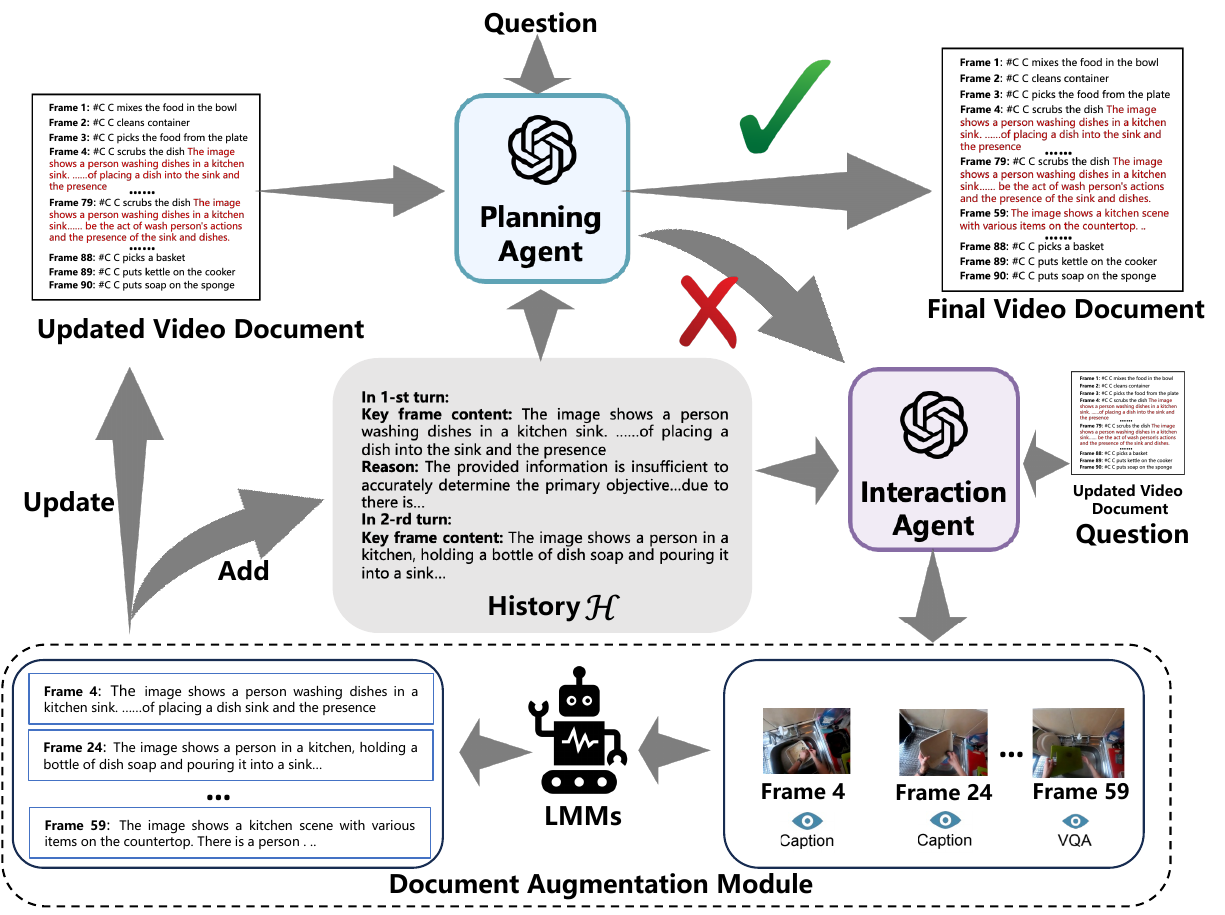}
\caption{\textbf{Illustration of the multi-stage agent interaction loop and answering module.} There are two agents in the multi-stage agent interaction loop: a planning agent to plan the next step and an interaction agent to dynamically find missing information and interact with the document augmentation module.}
    \label{fig2}
\end{figure}

Apart from the retrieved key frames, 
we further introduce a multi-stage agent interaction loop to dynamically find other potential key frames and augment them with different types of information for the final answer. We employ two distinct agents: planning and interaction agents, each tailored for a specific task. Both agents utilize the same LLM, \textit{i.e.,} GPT-3.5, but are given different prompts and reason varying types of information. The detail is shown in Fig. \ref{fig2}
\par
\textbf{Planning Agent.} Given the question $Q$ and the updated video document $\mathcal{AD}_i$ (where $i$ denotes the iteration step), along with the analysis history $\mathcal{H}$ from all previous steps (initialized as empty $\{\}$), the planning agent first determines whether the updated video document is sufficient to generate a confident answer. If the information is sufficient, the process moves to the answering module to get the final answer. If not, 
the agent provides an analysis of why the updated video document $\mathcal{AD}_i$ is inadequate. The agent then updates $\mathcal{H}_i$ with its analysis and passes the updated $\mathcal{H}_{i}$ to the interaction agent.\par
\textbf{Interaction Agent.} Given the current video document $\mathcal{AD}_i$ and the updated $\mathcal{H}_{i}$, the interaction agent finds $N$ frames with missing details crucial for answering $Q$:
\begin{gather}
    \resizebox{0.8\linewidth}{!}{$\forall n \in N, \hspace{0.1em} n \in T, \hspace{0.1em} n \notin \text{topk\_doc}, \hspace{0.1em} \text{and} \hspace{0.1em} N < K$}.
\end{gather}

In addition to identifying the missing $N$ key frames, the interaction agent also determines the augmented information type for each $ n \in N$. To achieve it, we give a task-specific prompt to the interaction agent, \textit{i.e., Your task is to determine which frame needs which type of information and can answer this question accurately, reasonably, and without contradiction. The two types of information are as follows: A: Given an image, get a detailed description of the image (image caption, just like what is shown in this image?) B: Given an image, get a response to the above question (visual question answering)}. 
After obtaining $N$ frames along with their required information type, the agent interacts with the document augmentation module to enrich the information of these frames to update the current video document for the next step.

\subsection{Answering Module} Given the final video document $\mathcal{AD}_{final}$, 
we employ another agent to provide a prediction using a chain-of-thought (CoT) approach~\cite{wei2022chain}. The answering module outputs the corresponding answer, the confidence score, and the reasoning behind the answer. Besides improving prediction accuracy, the CoT approach allows us to trace inference steps, ensuring transparency and explainability in the decision-making process. 

%% file: sec/4_experiment.tex
\section{Experiments}
In this section, we first introduce the datasets and implementation details and then present the results and ablations of our proposed DrVideo.

\subsection{Datasets and Metrics}
In our experiments, we evaluate our DrVideo using three well-established datasets, emphasizing its zero-shot long video understanding capabilities.
\par
\textbf{EgoSchema.} EgoSchema dataset \cite{mangalam2023egoschema} consists of 5000 multiple-choice questions sourced from 5000 three-minute egocentric videos. This dataset only has a subset of 500 questions with publicly accessible labels, while the full set is evaluated on the leaderboard.

\par
\textbf{MovieChat-1K.} MovieChat-1K \cite{song2023moviechat} is a longer video understanding benchmark, which contains 1000 videos from movies and TV shows, each video lasting approximately 10 minutes. This dataset has two modes: global mode and breakpoint mode. The global mode refers to analyzing the entire video to understand its overall content and context, while the breakpoint mode focuses on analyzing specific frames or scenes.
\par
\textbf{Video-MME.} Video-MME \cite{fu2024video} is a recently proposed benchmark for comprehensive video analysis evaluation. We evaluate DrVideo on the "long-term videos" split of this dataset (long split), where video lengths vary from 30 to 60 minutes, with an average duration of 44 minutes.\par

\begin{table}[t]
\centering
\small
\renewcommand{\arraystretch}{1}
\resizebox{0.85\linewidth}{!}{
\begin{tabular}{l| c| c c}
\toprule
Method  & (M)LLM & Subset & Fullset \\
\midrule
MoReVQA~\cite{min2024morevqa}\! & PaLM-2 & - & 51.7 \\
Vamos \cite{wang2023vamos}& GPT-4&51.2&48.3\\
ProViQ \cite{choudhury2023zero} &GPT-3.5 &57.1&-\\
LLoVi~\cite{LLoVi}\! & GPT-3.5 & 57.6 & 50.3 \\
IG-VLM~\cite{kim2024image}\! & GPT-4V & 59.8 & - \\
VideoAgent~\cite{wang2024videoagent}\! & GPT-4 & 60.2 & 54.1 \\
MVU \cite{ranasinghe2024understanding} & Mistral-13B&60.3 &37.6\\
LLoVi~\cite{LLoVi}\! & GPT-4 & 61.2 & - \\
VideoAgent~\cite{fan2024videoagent}\! & GPT-4 & \underline{62.8} & \underline{60.2}\\
\midrule
\textbf{DrVideo (ours)} & GPT-3.5 & \textbf{62.6} & - \\
\textbf{DrVideo (ours)} & GPT-4 & \textbf{66.4} & \textbf{61.0} \\
\bottomrule
\end{tabular}
}
\caption{Results on EgoSchema compared to existing LLM-based state-of-the-art methods.}
\label{tab:1}
\end{table}

Since EgoSchema and Video-MME are multi-choice tasks, we use accuracy as the evaluation metric for both datasets. The MovieChat-1K benchmark focuses on open-ended questions, We use GPT-assisted evaluation to assess both the accuracy (true/false) and quality (score 0-5) of the models. We select Gemini-Pro \cite{team2023gemini} as the evaluation assistant and use the same prompt \cite{videoChatGpt} to conduct a fair comparison.\par 
\begin{table}[t]
\centering
\resizebox{\columnwidth}{!}{
\begin{tabular}{l|cc|cc}
\toprule
\multirow{2}{*}{Method} & \multicolumn{2}{c}{Global} & \multicolumn{2}{c}{Breakpoint}  \\
 & Acc. & \phantom{ab}Score\phantom{ab} & Acc. & \phantom{ab}Score\phantom{ab} \\
\midrule
\multicolumn{5}{c}{\textit{Open-Source MLLM}} \\
\midrule
Video LLaMA~\cite{video-llama}&  51.7& 2.72 & 39.1 & 2.11 \\
Video Chat~\cite{li2023videochat} & 57.8 & 3.08 &  46.1& 2.32 \\
Video-ChatGPT~\cite{maaz2023video}& 47.6 & 2.89 & 48.0 & 2.43 \\ 
MovieChat~\cite{song2023moviechat} & 68.3&  3.15&  48.3& 2.46\\
MovieChat+~\cite{song2024moviechat+} & 71.2&  \underline{3.51}&  \underline{49.6}& \underline{2.62}\\
\midrule
\multicolumn{5}{c}{\textit{Based on Open-source Captioners and Proprietary LLMs}} \\
\midrule
MM-VID~\cite{lin2023mm}  & 58.6 & 2.86 & 10.4 & 0.56  \\
LLoVi~\cite{LLoVi} & 58.3 & 2.87 & 17.8 & 1.03 \\
VideoAgent\textsuperscript{*}~\cite{wang2024videoagent} & 65.4\textsuperscript{*}& 3.12\textsuperscript{*} & 31.6\textsuperscript{*} & 2.05\textsuperscript{*} \\
Sullam Jeoung, et al \cite{jeoung2024adaptive}& \underline{84.8} & - & - & - \\

\midrule
\textbf{DrVideo (ours)} & \textbf{93.1}&  \textbf{4.41}&  \textbf{56.4}& \textbf{2.75}  \\
\bottomrule
\end{tabular}
}
\caption{\small Performance comparison on the MoiveChat-1K~\cite{song2023moviechat} benchmark against state-of-the-art methods. \textsuperscript{*} represents re-implemented results, implementation details refer to appendix.}
\label{tab:2}
\end{table}
\subsection{Implementation Details}
For EgoSchema \cite{mangalam2023egoschema} dataset, we choose LaViLa \cite{zhao2023learning} as the captioning model to convert videos into documents. Note that the training data of our LaViLa model does not include EgoSchema videos, which is the same as LLoVi \cite{LLoVi}. For MovieChat-1K and Video-MME benchmarks, LLaVA-NeXT \cite{liu2024llavanext} is used as our captioning model to generate brief descriptions for each frame. We preprocess videos by simply sampling them at 0.5 FPS for EgoSchema \cite{mangalam2023egoschema} and MovieChat-1K \cite{song2023moviechat} and 0.2 FPS for Video-MME \cite{fu2024video}. LLaVA-NeXT \cite{liu2024llavanext} is chosen to augment key frames in both datasets via different prompts. In the comparison experiments, GPT-4 \cite{openai2023gpt4}, \ie, gpt-4-1106-preview, is used as the agent to evaluate the performance of our DrVideo on EgoSchema and MovieChat-1K. DeepSeek \cite{guo2024deepseek}, \ie, DeepSeek V2.5, is used as the agent to evaluate the performance on Video-MME. For the ablation study, we use GPT-3.5 \cite{gpt3.5}, \ie, gpt-3.5-turbo-1106, to evaluate the effectiveness of our DrVideo due to the API cost.

\subsection{Main Results}
\noindent\textbf{Comparative results on EgoSchema.} Table \ref{tab:1} compares our DrVideo with existing LLM-based state-of-the-art methods \cite{min2024morevqa, wang2023vamos, choudhury2023zero, LLoVi, kim2024image, wang2024videoagent, ranasinghe2024understanding, fan2024videoagent} on EgoSchema benchmark \cite{mangalam2023egoschema}. Specifically, compared with LLoVi \cite{LLoVi} and VideoAgent \cite{wang2024videoagent} that leverage the same visual captioner (LaViLa \cite{zhao2023learning}) and LLM (GPT-4), DrVideo significantly outperforms these methods by 5.2\% and 6.2\% respectively. Compared with VideoAgent \cite{fan2024videoagent} which uses video-specific models (ViCLIP from InternVid \cite{wang2023internvid} and Video-LLaVA \cite{videoLlava}), DrVideo still performs better and achieves 3.6\% higher accuracy on the subset evaluation. Moreover, DrVideo significantly outperforms IG-VLM \cite{kim2024image} by 6.6\%, which uses strong MLLM, \textit{i.e.,} GPT-4V. These results confirm our idea of converting the long-video understanding task into a long-document understanding task is feasible and our DrVideo 
can find the question-related key frames more accurately. 


\begin{table}[t]
\centering
\small
\renewcommand{\arraystretch}{1}
\resizebox{0.95\linewidth}{!}{
\begin{tabular}{l |c| c c}
\toprule
Method  & Frames & w/o Subs & w Subs \\
\midrule
\multicolumn{4}{c}{\textit{Proprietary MLLM}} 
\\
\midrule
Claude 3.5 Sonnet \cite{anthropic2024claude} & 20 & 51.2 & 54.7 \\
GPT-4V \cite{openai2023gpt4v}& 10 & 53.5 & 56.9\\
GPT-4o mini \cite{openai2024gpt4o}& 250 & 58.6&63.4 \\
Gemini 1.5 Flash \cite{team2024gemini}& 0.5 FPS & 61.1&68.8 \\
GPT-4o \cite{openai2024gpt4o}& 384 & 65.3	&72.1 \\
Gemini 1.5 Pro \cite{team2024gemini}& 0.5 FPS & \textbf{67.4}&\textbf{77.4} \\
\midrule
\multicolumn{4}{c}{\textit{Open-Source MLLM}} \\
\midrule
LongVA \cite{zhang2024long}&  128 & 46.2& 47.6  \\
VITA-8x7B \cite{fu2024vita}& 32 & 48.6	& 50.9 \\  
LLaVA-OneVision-72B \cite{li2024llava}& 32 & 60.0&	62.4  \\
Qwen2-VL-72B \cite{bai2023qwen}& 768 & 62.2&	74.3  \\
\midrule
\multicolumn{4}{c}{\textit{Based on Open-source Captioners and Proprietary LLMs}} \\
\midrule
VideoAgent\textsuperscript{*} ~\cite{wang2024videoagent}\! & 7.6 & 40.2\textsuperscript{*}  & 44.4\textsuperscript{*}  \\
LLoVi\textsuperscript{*} ~\cite{LLoVi}\! & 0.2 FPS & 45.4\textsuperscript{*}  & 67.7\textsuperscript{*}  \\
\midrule
\textbf{DrVideo (ours)} & 0.2 FPS & 51.7 & 71.7 \\
-only subs & 0.2 FPS & - & 68.5 \\
\bottomrule
\end{tabular}
}
\caption{Video-MME long split results. \textit{w/o subs} represents the model without any subtitles. \textit{w subs} represents that the subtitles corresponding to the sampled frames are both fed into the LLM to evaluate the performance. Our
DrVideo outperforms a strong proprietary MLLM (Claude 3.5 Sonnet) and some open-source MLLMs (\textit{e.g.,} VITA-8x7B). \textsuperscript{*} represents re-implemented results, implementation details refer to appendix.}

\label{tab:3}
\end{table}

\noindent\textbf{Comparative results on longer video benchmarks.} 
To further emphasize the advantage of our method on longer videos, we evaluate our DrVideo on the other two video benchmarks, \textit{i.e.,} MoiveChat-1K (10 minutes) \cite{song2023moviechat}, and the long split of Video-MME (average 44 minutes) \cite{fu2024video}. Table \ref{tab:2} presents the comparative results of our DrVideo and the current state-of-the-art methods \cite{song2023moviechat,li2023videochat,video-llama,videoChatGpt, LLoVi} on the MovieChat-1K benchmark. Table \ref{tab:3} shows the comparative results of our DrVideo on the Video-MME benchmark, with comparisons against three types of methods: proprietary MLLMs \cite{anthropic2024claude, openai2023gpt4v, openai2024gpt4o, team2024gemini}, open-source MLLMs \cite{zhang2024long,fu2024vita,li2024llava,bai2023qwen}, and methods based on open-source captioners and proprietary LLMs \cite{LLoVi, wang2024videoagent}. 

Compared to methods like LLoVi \cite{LLoVi} and VideoAgent \cite{wang2024videoagent} that are based on open-source captioners (\textit{e.g.,} LLaVA-NeXT \cite{liu2024llavanext}) and proprietary LLMs (\textit{e.g.,} GPT-4 \cite{openai2023gpt4}), DrVideo achieves a 8.3\% improvement in global mode and a 24.8\% improvement in breakpoint mode on the MovieChat-1K benchmark in Table  \ref{tab:2}. It also achieves at least 6.3\% improvement under the w/o subtitle setting on the Video-MME benchmark in Table \ref{tab:3}. This indicates that DrVideo can accurately identify question-related frames and avoid critical information loss through augmentation, highlighting the potential of document retrieval methods for processing longer videos. Compared to proprietary MLLMs and open-source MLLMs, DrVideo outperforms Claude 3.5 Sonnet \cite{anthropic2024claude} by 0.5\% and some open-source MLLMs (\textit{e.g.,} VITA-8x7B \cite{fu2024vita}) under the w/o subtitle setting on Video-MME benchmark. It also achieves a 21.9\% improvement in global mode and a 6.8\% improvement in breakpoint mode on MovieChat-1K benchmark. 
Moreover, combining the subtitle on the Video-MME benchmark, DrVideo achieves 71.7\% accuracy and outperforms many heavily-engineered large-scale proprietary models (\textit{e.g.,} Gemini 1.5 Flash \cite{team2024gemini}, GPT-4o mini \cite{openai2024gpt4o}, and GPT-4V \cite{openai2023gpt4v}). To better understand this significant improvement, we conduct an experiment where only the subtitle, question and options are input into the LLM (\textit{i.e.,} DeepSeek \cite{guo2024deepseek}) to predict the answer and the accuracy is 68.5\%. This suggests: (i) subtitles contain extensive question-related information, playing a crucial role in understanding long videos; (ii) compared with LLoVi and VideoAgent, DrVideo enhances visual information via document retrieval and augmentation, effectively integrating subtitles and resulting in a 3.2\% improvement.

\subsection{Ablation Studies}\label{experiment:ablation}
Unless otherwise specified, we use GPT-3.5 as the default setting in the below experiments on EgoSchema subset \cite{mangalam2023egoschema}.\par

\newcommand{\cmark}{\ding{51}}%
\newcommand{\xmark}{\ding{55}}%
\begin{table}[t]

\centering
\setlength{\tabcolsep}{20pt}
\renewcommand{\arraystretch}{1.3}
\resizebox{0.85\linewidth}{!}{
\begin{tabular}{c c c |c}
\toprule
RM & MSAIL & CoT & Acc.(\%) \\
\midrule
\cmark & \cmark & \cmark & \textbf{62.6} \\
\cmark & \cmark & \xmark & 62.2 \\
\cmark & \xmark & \cmark & 60.6 \\
\xmark & \cmark & \cmark & 59.4 \\ 
\xmark & \xmark & \cmark & 57.4 \\
\bottomrule
\end{tabular}
}
\caption{Ablation results on different combinations of the retrieval module (RM), the multi-stage agent interaction loop (MSAIL), and the CoT on EgoSchema. }
\label{tab4}
\end{table}

\begin{table}[t]
    \centering
    \small
    \renewcommand{\arraystretch}{1.3}  
    \setlength{\tabcolsep}{9pt}  
    \resizebox{\linewidth}{!}{  
        \begin{tabular}{cc}
            \begin{minipage}[t]{0.45\linewidth}
                \centering
                \begin{tabular}{c c | c}
                \toprule
                VQA & Caption & Acc. (\%) \\
                \midrule
                \checkmark & \checkmark & \textbf{62.6} \\
                \texttimes & \checkmark & 60.4 \\
                \checkmark & \texttimes & 61.8 \\
                \bottomrule
                \end{tabular}
                \subcaption{Performance with different types of information}
                \label{tab5_1}
            \end{minipage} 
            &
            \begin{minipage}[t]{0.55\linewidth}
                \centering
                \begin{tabular}{c | c}
                \toprule
                Top-K & Acc. (\%) \\
                \midrule
                5 & \textbf{62.6} \\
                10 & 61.4 \\
                20 & 60.6 \\
                \bottomrule
                \end{tabular}
                \subcaption{Performance with different Top-$K$ frames}
                \label{tab5_2}
            \end{minipage}
        \end{tabular}
    }
    \caption{Ablation results with different settings on EgoSchema.} 
    \label{tab5}
\end{table}
\noindent\textbf{Effects of different components of DrVideo. } 
We conduct experiments with different combinations of the individual components in DrVideo, including the retrieval module, the multi-stage agent interaction loop, and the CoT used in the answering module. The results are presented in Table \ref{tab4}. We can see that: (i) without the retrieval module and the agent loop, the performance of our DrVideo is similar to LLoVi~\cite{LLoVi}; 
(ii) By adding the retrieval module and then the agent loop, the performance of DrVideo improves from 57.4\% to 
60.6\% and then to 62.6\%. This demonstrates that the retrieval module can identify key frames via measuring semantic similarity, while the agent loop can locate additional key frames through contextual inference; (iii) the additional CoT contributes to the accuracy improvement from 62.2\% to 62.6\%, demonstrating its effectiveness. 

\noindent\textbf{Effects of different types of augment information.}
To better enhance the required information for each potential key frame, we define two types of augment information, namely VQA (question-related information) and caption (general detailed information). Table \ref{tab5_1} presents the performance of DrVideo with different types of information. We can see that when only one type of information (VQA or caption) is used for augmentation, the model's performance declines. This suggests the necessity of adaptively augmenting different types of information for each potential key frame. 
\par
\noindent\textbf{Effects of different initial top $K$ key frames.} By default, the number of the initial key frames is set as $K=5$. Intuitively, more frames should lead to better performance at the cost of increasing the complexity. 
However, our ablation in Table \ref{tab5_2} shows an opposite trend, \ie, more key frames resulting in worse performance. 
This suggests that enhancing more frames is not necessarily beneficial, as it can introduce a large amount of noisy and irrelevant information that interferes with the LLM's judgment. 

\begin{figure}[t]
    \centering
    \includegraphics[width=0.95\linewidth]{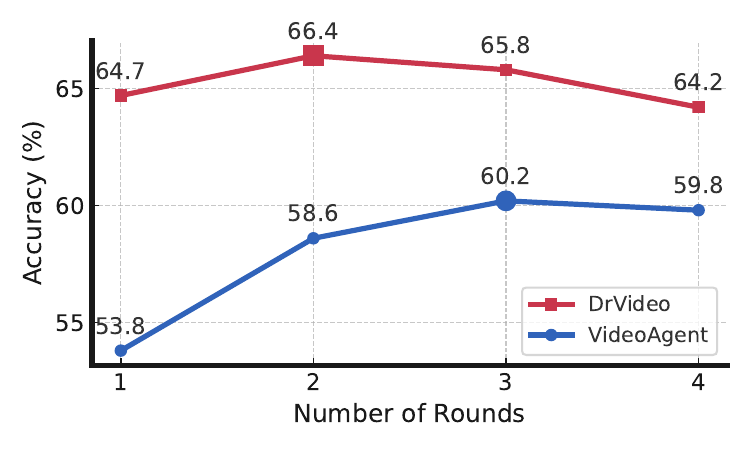}
\caption{Performance of different rounds with DrVideo and VideoAgent \cite{wang2024videoagent} on EgoSchema. To align with VideoAgent, GPT-4, \textit{i.e.,} gpt-4-turbo-1106-preview, is used as the LLM agents.}
    \label{fig3}
\end{figure}
\begin{table}[t]
    \centering
    \renewcommand{\arraystretch}{1.3}
    \setlength{\tabcolsep}{20pt} 
    \resizebox{0.90\linewidth}{!}{
        \begin{tabular}{c| c |c}
            \toprule
            Captioning Model & Type &Acc.(\%) \\
            \midrule
            LaViLa \cite{zhao2023learning} &  Clip-based& \textbf{62.6}\\
            LLaVA-NeXT \cite{liu2024llavanext} &  Frame-based& 61.2\\
            BLIP-2 \cite{li2023blip2} &  Frame-based& 59.6\\
            \bottomrule
        \end{tabular}
    }
    \caption{Performance of different VLMs on EgoSchema.}
    \label{tab6}
\end{table}
\noindent\textbf{Effects of iterative rounds.} We also examined the impact of the iterative rounds $I$ on model performance. Fig. \ref{fig3} shows the performance of different iterative rounds with DrVideo and VideoAgent \cite{wang2024videoagent}. We have the following observations. (i) DrVideo consistently outperforms VideoAgent, reaching the peak performance at $I=2$, while VideoAgent peaks at 
$I=3$. This indicates that DrVideo can locate the information needed to answer questions in fewer interactions and with higher quality, underscoring the superiority of our document retrieval-based framework. (ii) When the number of interactions exceeds two, DrVideo’s performance declines. This is likely due to more noisy and irrelevant information being introduced into the LLM with the increase of the iterations. This further confirms that excessive information is not necessarily beneficial. 

\begin{table}[t]
    \centering
    \renewcommand{\arraystretch}{1.3}
    \setlength{\tabcolsep}{20pt} 
    \resizebox{0.90\linewidth}{!}{
        \begin{tabular}{c |c| c|c}
            \toprule
            LLMs & Type& Size &Acc.(\%) \\
            \midrule
            Mistral-8x7B \cite{jiang2023mistral} &  Open-Source&8x7B &47.6\\
            DeepSeek \cite{guo2024deepseek} &  Proprietary& N/A &61.2\\
            GPT-3.5 \cite{gpt3.5} &  Proprietary& N/A &62.6\\
            GPT-4 \cite{openai2023gpt4} &  Proprietary& N/A &\textbf{66.4}\\

            \bottomrule
        \end{tabular}
    }
    \caption{Performance of different LLMs on EgoSchema.}
    \label{tab7}
\end{table}
\begin{figure*}[t]
    \centering
    \includegraphics[scale=0.55]{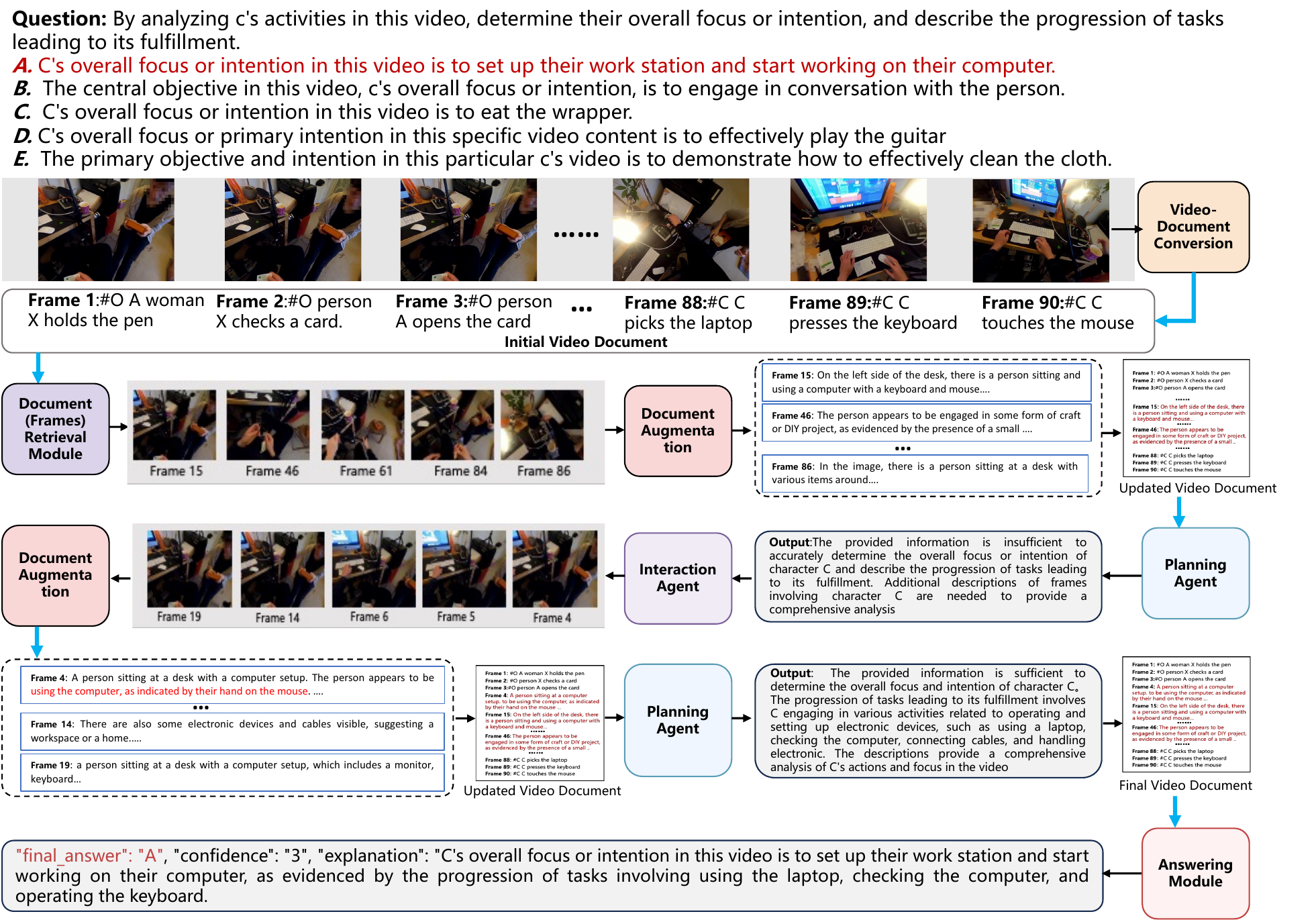}

    \caption{Case study on an instance from EgoSchema. DrVideo accurately identifies key frames and chooses the correct answer.}
    \label{fig4}
\end{figure*}

\noindent\textbf{Effects of different foundation models.} DrVideo incorporates two types of foundational models — large language model (LLM), and visual language model (VLM), {where VLM is used as the frame captioner while LLM is used as the planning agent, the interaction agent, and the answering agent.} 
To assess the impact of the captions produced by different VLMs, we examine three state-of-the-art VLMs: frame-based BLIP-2 \cite{li2023blip2}, LLaVA-NeXT \cite{liu2024llavanext}, and clip-based LaViLa \cite{zhao2023learning}, as presented in Table \ref{tab6}. We observe that LaViLa achieves the best performance, while BLIP-2 is the worst. 
It indicates the superiority of the clip-based model which can capture motion information in videos and is more suited for video understanding.

To evaluate the effectiveness of DrVideo with different LLMs, Table \ref{tab7} shows the results of DrVideo using open-source LLM (Mistral-8x7B \cite{jiang2023mistral}) or Proprietary LLM (DeepSeek \cite{guo2024deepseek}, GPT-3.5 \cite{gpt3.5}, or GPT-4 \cite{openai2023gpt4}) as the planning, interaction and answering agents. 
From the results, we can see that: (i) different LLMs in DrVideo exhibit significant performance gaps and the proprietary LLM significantly outperforms the open-source LLM, indicating DrVideo’s dependency on the choice of LLM; (ii) as the LLM's capability increases, the performance of DrVideo improves, demonstrating the great potential of DrVideo. It’s worth noting that our main contribution is the introduction of a framework for translating the long video understanding into a document retrieval and understanding task to effectively leverage the power of large language models, rather than the employment of any specific models. Our DrVideo can be further improved by combining with better LLM and VLM models (\textit{e.g.,} tarsier \cite{wang2024tarsier}).

\noindent\textbf{Case study.} Here we give a detailed example from EgoSchema \cite{mangalam2023egoschema} in Fig. \ref{fig4} to qualitatively demonstrate the capability of our DrVideo. 
Specifically, the question is to determine the overall focus or intention and describe the progression of tasks leading to its fulfillment. DrVideo first converts the raw video into an initial video document, which is then processed by the frame retrieval module to obtain key frames (frames 15, 46, 61, 84, and 86) and is augmented with detailed key frame descriptions. The updated video document is then fed into the planning agent to assess if sufficient information is available for answering, \ie, the provided information is sufficient to infer character C’s main focus or intent. This updated document, along with the rationale, is input to the interaction agent to identify the additional key frames (frames 4, 5, 6, 14, and 19 here). After further augmentation, the updated document is sent back to the planning agent, for which the agent confirms its sufficiency. Finally, the answering module takes in the final video document and predicts the correct answer.

\section{Conclusion}
We have proposed DrVideo, a document retrieval-based system for long video understanding. Different from previous LLM-based methods, DrVideo proposes to adapt long-video understanding to long-document understanding and searches for missing information via document retrieval and multi-stage agent loop. Extensive comparative experiments and ablation studies on the three challenging datasets, \textit{i.e.,} EgoSchema, MovieChat-1K, and Video-MME, demonstrate the effectiveness of DrVideo. We believe our work can serve as a strong baseline to stimulate further developments on the topic of long video understanding.

\section{Limitation}
Although Drvideo achieves impressive results on different long video benchmarks, it has several limitations. 
(i) The longest video DrVideo can handle depends on the maximum token length of the LLM, which is the bottleneck for much longer videos than the existing benchmarks, \textit{e.g.,} 10 hours.
(ii) It has space for further improvement on how to generate sufficient information while minimizing irrelevant information in the video document. 


%% file: sec/X_suppl.tex
\clearpage
\setcounter{page}{1}
\maketitlesupplementary

\section{More Implementation Details}
\vspace{-0.25cm}

\textbf{\textit{Experiments Compute Resources.}} All experiments are conducted on single NVIDIA RTX 4090 GPU. The minimal GPU memory requirement is 24GB. We set the temperature to 0 for all experiments using GPT-3.5 \cite{gpt3.5}, GPT-4 \cite{openai2023gpt4}, and DeepSeek \cite{guo2024deepseek}. \par
\textbf{\textit{Prompt Details.}} We provide detailed prompts for all agents (planning agent, interaction agent, and answering agent) in the EgoSchema benchmark \cite{mangalam2023egoschema}. Planing agent is to determine whether the video captions are sufficient for answering the question. Below is the planning agent prompt: \par
\begin{tcolorbox}[
    colback=white!95!black,    
    colframe=black,            
    width=\linewidth,          
    sharp corners,             
    boxrule=0.5mm,             
    coltitle=black             
]
\textbf{User}

You are given some language descriptions of a first-person view video along with a question about the video. 

1.The video is 3 minutes long, containing a total of 90 frames.

2. Each sentence in these language descriptions represents the text description for a single frame.

3. The format of each sentence is {frame id, description}. The frame id indicates the temporal position of the frame, ranging from 1 to 90.

Here are the original descriptions of this video: \textcolor{blue}{Documents}

Here is the question: \textcolor{blue}{Question}

Here is the memory: \textcolor{blue}{Memory}

Your task is to determine whether these descriptions above can answer the question accurately, reasonably, and without contradiction. 

If your answer is yes, please give me an reasonable explanation. the output will be as follows: \textbf{\{``confidence": ``1", ``explanation": [``xxxx"]\}}

If your answer is no, the confidence is 0, indicating the provided information is insufficient. Please give me a reasonable explanation for what frame is missing. For each frame identified as potentially relevant, provide a concise description focusing on essential visual elements(e.g., objects, humans, interactions, actions, and scenes) in the explanation. The output will be as follows: \textbf{\{``confidence": ``0", ``explanation": [``xxxx"]\}}

You must not provide any other response or explanation.

\rule{\linewidth}{0.4pt}

\textbf{Assistant}

\textcolor{blue}{{``confidence": ``0/1", ``explanation": [``xxxx"]}}
\end{tcolorbox}

Interaction agent is used to find potential missing key frames with different types of information. The interaction agent prompt in the EgoSchema benchmark is shown as below:\par

\begin{tcolorbox}[
    colback=white!95!black,    
    colframe=black,            
    width=\linewidth,          
    sharp corners,             
    boxrule=0.5mm,             
    coltitle=black             
]
\textbf{User}

You are given some language descriptions of a first-person view video along with a question about the video. 

1.The video is 3 minutes long, containing a total of 90 frames.

2. Each sentence in these language descriptions represents the text description for a single frame.

3. The format of each sentence is {frame id, description}. The frame id indicates the temporal position of the frame, ranging from 1 to 90.

Here are the original descriptions of this video: \textcolor{blue}{Documents}

Here are the memory: \textcolor{blue}{Memory}

To answer the following question: \textcolor{blue}{Question}

Theses descriptions are insufficient and cannot answer this question accurately, reasonably, and without contradiction.

Your task is to determine which frame needs which type of information and can answer this question accurately, reasonably, and without contradiction. 

The two types of information are as follows:

A: Given an image, get a detailed description of the image (image caption, just like what is shown in this image?)

B: Given an image, get a response to the above question (visual question answering)

Please note that frame selections range from 1 to 90. These frames (\textcolor{blue}{type\_A}) already have type A information and these frames (\textcolor{blue}{type\_B}) already have type B information, please note not to repeatedly select this type of information from these frames. Please note that the the key of frame only one number. The output must be as follows:
\textbf{[{``frame": ``1/2/3/.../90", ``type": ``A/B"}]}

\rule{\linewidth}{0.4pt}

\textbf{Assistant}

\textcolor{blue}{[{``frame": ``1/2/3/.../90", ``type": ``A/B"}]}
\end{tcolorbox}

Finally, the answering agent is used to predict the answer once the video captions are sufficient. The answering agent is shown as below:\par

\begin{tcolorbox}[
    colback=white!95!black,    
    colframe=black,            
    width=\linewidth,          
    sharp corners,             
    boxrule=0.5mm,             
    coltitle=black             
]
\textbf{User}

You are individual C, with others represented as O. Your task is to answer a question related to this video, choosing the correct option out of five possible answers. You are given some language descriptions of a first person view video along with a question about the
video. 

1.The video is 3 minutes long, containing a total of 90 frames.

2. Each sentence in these language descriptions represents the text description for a single frame.

3. The format of each sentence is {frame id, description}. The frame id indicates the temporal position of the frame, ranging from 1 to 90.

Here are the descriptions of this video: \textcolor{blue}{Documents}

Please answer the following question: \textcolor{blue}{Question}

Here are the choices. A: \textcolor{blue}{option1} B: \textcolor{blue}{option2} C: \textcolor{blue}{option3} D: \textcolor{blue}{option4} E: \textcolor{blue}{option5}

The question has 5 choices, labeled as A, B, C, D, E. Please think step by step and write the best answer index. Note your final answer must be one of the letters (A, B, C, D, or E), the confidence must be one of the letters (1, 2, 3), please provide a concise one-sentence explanation for your chosen answer. the output must be the following format. You must not provide any other response or explanation. 

\textbf{\{``final\_answer": ``xxx", ``confidence": ``xxx", ``explaination": ``xxx"\}}

\rule{\linewidth}{0.4pt}

\textbf{Assistant}

\textcolor{blue}{\{``final\_answer": ``xxx", ``confidence": ``xxx", ``explaination": ``xxx"\}}
\end{tcolorbox}

\textbf{\textit{Details of LaViLa.}} For the experiments on EgoSchema \cite{mangalam2023egoschema}, we utilize LaViLa \cite{zhao2023learning} as the captioner, a CLIP-based captioning model. LaViLa processes input clips with a resolution of 4 × 336 × 336 and is trained on the Ego4D dataset \cite{grauman2022ego4d}. The original LaViLa training set contains 7,743 videos with 3.9 million video-text pairs, while the validation set includes 828 videos with 1.3 million video-text pairs. Since the EgoSchema dataset is cropped from Ego4D and designed for zero-shot evaluation, using the original LaViLa model could lead to overlap with EgoSchema videos, resulting in an unfair comparison. To address this, we use the retrained LaViLa model which same as LLoVi \cite{LLoVi} that do not have any overlap with EgoSchema videos to avoid unfair comparison with other methods. The checkpoints are available at \url{https://drive.google.com/file/d/1AZ5I4eTUAUBX31rL8jLB00vNrlFGWiv8/view}.\par
\textbf{\textit{Details of LLaVA-NeXT.}} For the experiments on MovieChat-1K \cite{song2023moviechat} and Video-MME \cite{fu2024video}, we utilize LLaVA-NeXT \cite{liu2024llavanext} as the captioner, which is a frame-based captioning model. LLaVA-NeXT used in our framework has 7B parameters and the checkpoints are available at \url{https://huggingface.co/llava-hf/llava-v1.6-mistral-7b-hf}. We also use the same LLaVA-NeXT model to augment key frames with different prompts in both datasets.\par

\textbf{\textit{VideoAgent Reproduce Details.}} To evaluate the performance of VideoAgent \cite{wang2024videoagent} on MovieChat-1K benchmark \cite{song2023moviechat} and make a fair comparison, we select the LLaVA-NeXT model as the captioner and preprocess videos by simply sampling them at 0.5 FPS. GPT-4 \cite{openai2023gpt4}, \textit{i.e.,} gpt-4-1106-preview, is used as the agent to predict answer, self reflect, and find missing information. The prompt for each agent is same as the original paper. The max interaction rounds is set to 3 and the initial sampled frames for whole video is set to 5, which is the default setting in the original paper. The EVA-CLIP-8B-plus model \cite{sun2024eva}, a state-of-the-art CLIP model that includes a vision encoder with 7.5 billion parameters and a text encoder with 0.7 billion parameters, is used for frame retrieval to align with the original paper \cite{wang2024videoagent}. The above setting is used to the global mode and breakpoint mode to ensure reproducibility. The checkpoints of LLaVA-NeXT model are available at \url{https://huggingface.co/llava-hf/llava-v1.6-mistral-7b-hf}. The website of GPT-4 API is available at \url{https://openai.com/index/gpt-4-api-general-availability/}. The checkpoints of EVA-CLIP-8B-plus model are available at \url{https://huggingface.co/BAAI/EVA-CLIP-8B-448}.\par
To evaluate the performance of VideoAgent \cite{wang2024videoagent} on Video-MME benchmark \cite{fu2024video} and make a fair comparison, we preprocess videos by simply sampling them at 0.2 FPS and DeepSeek, \textit{i.e.,} DeepSeek V2.5, is used as the agent to predict answer, self reflect, and find missing information. Except for those, the other settings are the same as the MovieChat-1K benchmark. The website of DeepSeek API is available at \url{https://api-docs.deepseek.com/}. Note that for the experiment with subtitles, we select the subtitles corresponding to the sampled frames to add to the LLM, rather than using all the subtitles. This is why the performance of VideoAgent is significantly lower than the other LLM-based methods under the \textit{w subs} setting. \par
\textbf{\textit{LLoVi Reproduce Details.}} To evaluate the performance of LLoVi \cite{LLoVi} on Video-MME benchmark and make a fair comparison, we select the same LLaVA-NeXT model \cite{liu2024llavanext} as the captioner and preprocess videos by simply sampling them at 0.2 FPS. DeepSeek \cite{guo2024deepseek}, \textit{i.e.,} DeepSeek V2.5, is used as the agent to summary the sampled captions and predict the final answer. The summaries words are set to 500 and the prompt of each prompt is same as the original paper \cite{LLoVi}. The website of DeepSeek API is available at \url{https://api-docs.deepseek.com/}. The checkpoints of LLaVA-NeXT model are available at \url{https://huggingface.co/llava-hf/llava-v1.6-mistral-7b-hf}. \par

\textbf{\textit{Only Subs Reproduce Details.}} In Section 4.3 of our paper, we conduct an experiment where only the subtitle, question and options are input into the LLM (\textit{i.e.,} DeepSeek \cite{guo2024deepseek}) to predict the answer. Specifically, the subtitles used in the LLM are aligned with the sampled frames, while we preprocess videos by simply sampling them at 0.2 FPS. The prompt for the LLM is shown in below:\par

\begin{tcolorbox}[
    colback=white!95!black,    
    colframe=black,            
    width=\linewidth,          
    sharp corners,             
    boxrule=0.5mm,             
    coltitle=black             
]
\textbf{User}

This video's subtitles are listed below: 

\textcolor{blue}{subtitle\_1}

\textcolor{blue}{subtitle\_2}

...

\textcolor{blue}{subtitle\_$n$}

Select the best answer to the following multiple-choice question based on the subtitles and summary. Respond with only the letter (A, B, C, or D) of the correct option.

\textcolor{blue}{Question}

\textcolor{blue}{OptionA}

\textcolor{blue}{OptionB}

\textcolor{blue}{OptionC}

\textcolor{blue}{OptionD}

\rule{\linewidth}{0.4pt}

\textbf{Assistant}

\textcolor{blue}{A/B/C/D}
\end{tcolorbox}
where $n$ represents the number of subtitles under the 0.2 FPS sampling.

\section{More Ablation Studies}
In this paper, we choose LaViLa \cite{zhao2023learning} as the captioning model to convert videos into documents. we preprocess videos by simply sampling them at 0.5 FPS for EgoSchema. To evaluate the impact of different sampling rates on the model's performance. We conduct the experiments with different sampling rates (1 FPS, 0.5 FPS, and 0.25 FPS) to convert videos into documents on EgoSchema subset. Besides, we use GPT-3.5 \cite{gpt3.5} as the planning agent, interaction agent, and answering agent for the below comparisons.\par

\begin{table}[t]
    \centering
    \renewcommand{\arraystretch}{1.3}
    \setlength{\tabcolsep}{20pt} 
    \resizebox{0.90\linewidth}{!}{
        \begin{tabular}{c| c }
            \toprule
            Sampling Rate & Accuracy (\%) \\
            \midrule
            1 FPS &  61.6\\
            0.5 FPS & \textbf{62.6}\\
            0.25 FPS &  58.8\\
            \bottomrule
        \end{tabular}
    }
    \caption{Performance of different sampling rate on EgoSchema.}
    \label{tab12}
\end{table}

\begin{algorithm}[t]
\caption{DrVideo}
\label{alg:drvideo}
\KwIn{$V$, $Q$}
\textbf{$\blacktriangleright$ Video-Document Conversion Module}\;
$\text{Doc}_{init} \gets \text{getDoc}(V, \phi_{vlm}, \mathcal{P})$\;
$\mathcal{E}_{\text{doc}} \gets \text{getEmbedding}(\text{Doc}_{init}, \phi_{emb})$\;

\textbf{$\blacktriangleright$ Document Retrieval Module}\;
Initialize $\mathcal{RT}$ as $Q$\;
$\mathcal{E}_{\mathcal{RT}} \gets \text{getEmbedding}(\mathcal{RT}, \phi_{emb})$\;
$\text{topk\_doc} \gets \text{Retrieval}(\mathcal{E}_{\mathcal{RT}}, \mathcal{E}_{\text{doc}}, K)$\;

\textbf{$\blacktriangleright$ Document Augmentation Module}\;
Initialize $\mathcal{AP}$\;
$\mathcal{AD}_0 \gets \text{Augment}(\text{topk\_doc}, \text{Doc}_{init}, \mathcal{AP})$\;
$\mathcal{H} \gets \text{addToMemory}(\text{topk\_doc})$\;

\textbf{$\blacktriangleright$ Multi-Stage Agent Interaction}\;
Initialize $i \gets 0$\;
\While{$i \leq I$}{
  \textbf{Planning Agent:}\;
  $\mathcal{S}, R \gets \text{checkSufficient}(\mathcal{AD}_i, \mathcal{H}_i, Q)$\;
  \If{$\mathcal{S} == 1$}{
    break\;
  }
  \Else{
    $\mathcal{H} \gets \text{addToMemory}(\mathcal{H}, R)$\;
    \textbf{Interaction Agent:}\;
    $\mathcal{M} \gets \text{FindMissInfo}(\mathcal{AD}_i, \mathcal{H}, Q)$\;
    $N, \mathcal{AP} \gets \mathcal{M}$\;
    $\mathcal{H} \gets \text{addToMemory}(\mathcal{H}, N)$\;
    $i \gets i + 1$\;
    $\mathcal{AD}_i \gets \text{Augment}(N, \mathcal{AD}_i, \mathcal{AP})$\;
  }
}
\textbf{$\blacktriangleright$ Answering Module}\;
$P \gets \text{GetAnswer}(\mathcal{AD}_i)$\;
\Return $P$\;
\end{algorithm}

\begin{figure*}[t]
    \centering
    \includegraphics[scale=0.55]{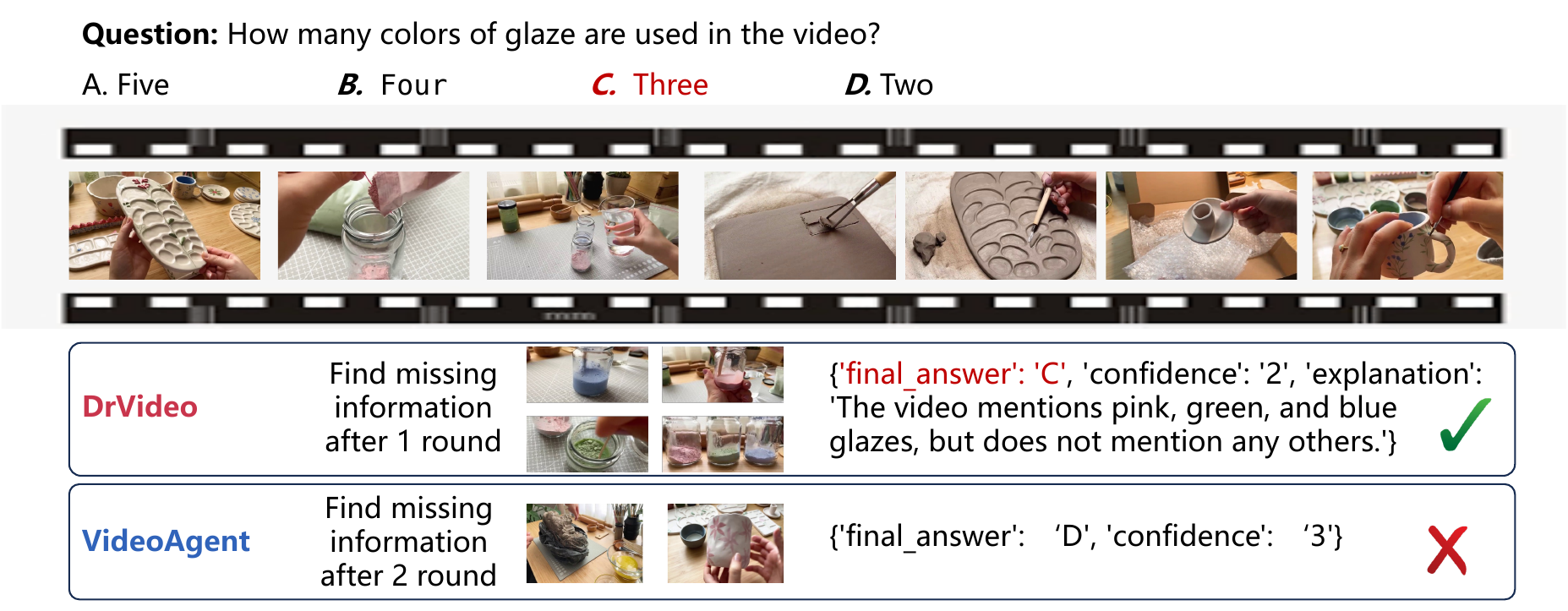}

    \caption{Case study on an instance from Video-MME. Long Case of DrVideo. This video contains 33 minutes.}
    \label{long_case}
\end{figure*}

\begin{figure*}[t]
    \centering
    \includegraphics[scale=0.55]{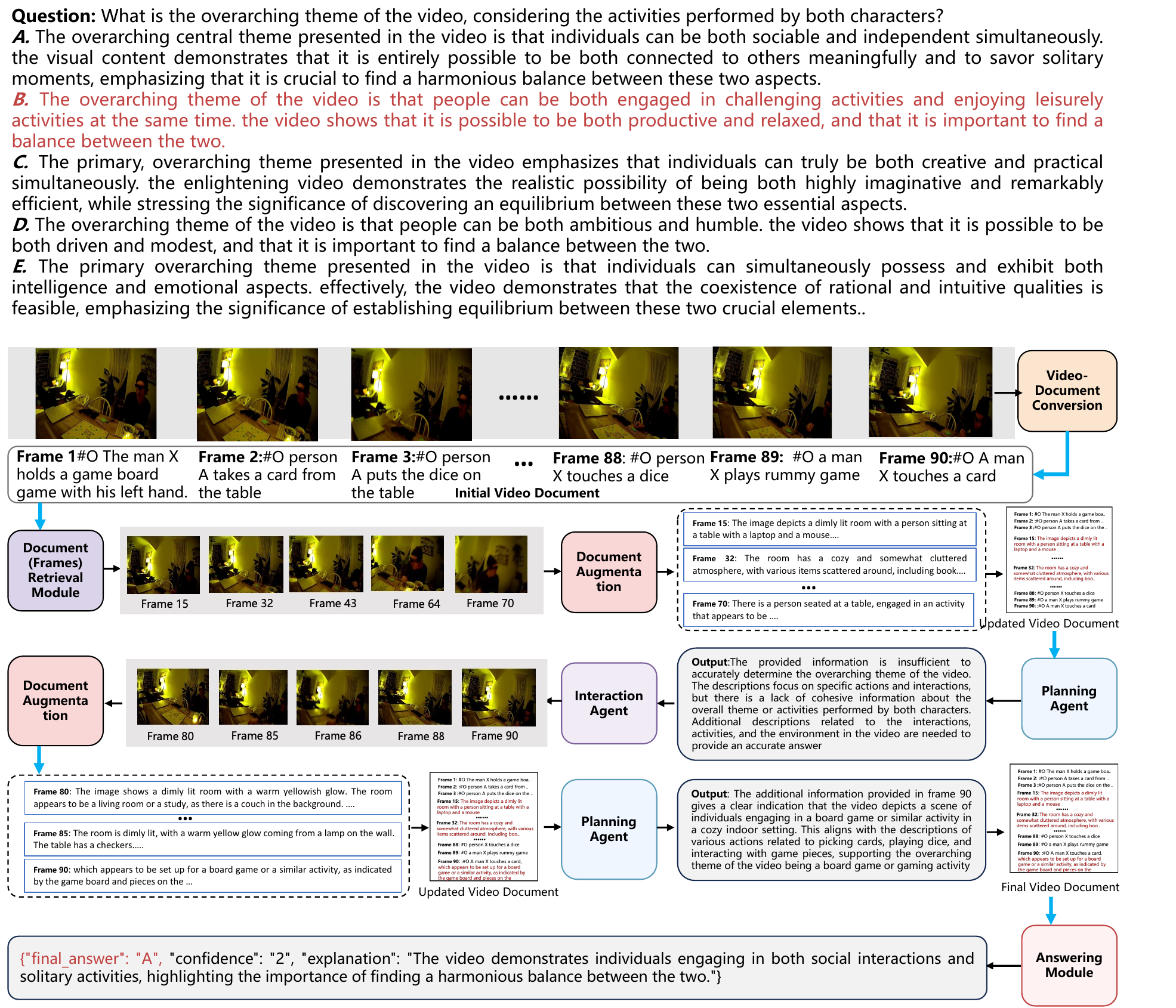}

    \caption{Case study on an instance from EgoSchema. Failure Case of DrVideo.}
    \label{fail_case}
\end{figure*}
Table \ref{tab12} presents the results and we have the following observations: (i) When the sampling rate is set to 1 FPS, the performance is lower than the sampling rate is set to 0.5 FPS. It indicates that a higher sampling frequency is not always better, as higher frequencies introduce more redundant information. (ii) When the sampling rate is set to 0.25 FPS, the performance is also lower than the sampling rate is set to 0.5 FPS. This suggests that a low sampling frequency leads to the loss of significant information, which in turn causes a decline in performance. (iii) When the sampling rate is set to 0.5 FPS, DrVideo achieves the best performance. This highlights the importance of selecting an appropriate sampling rate to reduce redundant information while retaining critical details for DrVideo. It also leaves room for future improvements by adaptively sampling the video based on the question.

\section{Detailed Algorithm}
In Algorithm \ref{alg:drvideo}, we present the algorithm behind DrVideo to give the reader a clearer understanding. The definition of the symbols has already been provided in the main text.

\section{More Case Studies}
Fig. \ref{long_case} shows how DrVideo can accurately solve hour-long
videos from the long split of Video-MME benchmark. The question is about figuring out the colors of glaze used in the video, which not only requires the model to have a comprehensive understanding of the video but also a clear understanding of its local details. DrVideo accurately  identifies the necessary information and predicts the answer correctly, outperforming state-of-the-art models like VideoAgent. This highlight the potential of our document retrieval method in handle longer videos.\par
Besides, we also provide a failure case to explore the limitations of our DrVideo as shown in Fig. \ref{fail_case}. The question is determine the overarching theme of the video, considering the activities performed by both characters. DrVideo, after undergoing key frame retrieval enhancement and multi-stage interaction loop continuation retrieval enhancement, evaluates whether the information collected so far can accurately answer the question. However, due to the inability of the VLM to accurately describe the video content, DrVideo makes incorrect judgments, ultimately leading to a wrong answer. This indicates that DrVideo heavily relies on the capabilities of both LLMs and VLMs. We believe DrVideo will be improved with the development of better LLMs and VLMs in the future.



%% file: arxiv.bbl
\begin{thebibliography}{77}
\providecommand{\natexlab}[1]{#1}
\providecommand{\url}[1]{\texttt{#1}}
\expandafter\ifx\csname urlstyle\endcsname\relax
  \providecommand{\doi}[1]{doi: #1}\else
  \providecommand{\doi}{doi: \begingroup \urlstyle{rm}\Url}\fi

\bibitem[Achiam et~al.(2023)Achiam, Adler, Agarwal, Ahmad, Akkaya, Aleman, Almeida, Altenschmidt, Altman, Anadkat, et~al.]{openai2023gpt4}
Josh Achiam, Steven Adler, Sandhini Agarwal, Lama Ahmad, Ilge Akkaya, Florencia~Leoni Aleman, Diogo Almeida, Janko Altenschmidt, Sam Altman, Shyamal Anadkat, et~al.
\newblock Gpt-4 technical report.
\newblock \emph{arXiv preprint arXiv:2303.08774}, 2023.

\bibitem[Anthropic(2024)]{anthropic2024claude}
Anthropic.
\newblock {Claude 3.5 Sonnet: Enhanced Conversational AI}.
\newblock \url{https://www.anthropic.com/index/claude-3.5-sonnet/}, 2024.

\bibitem[Bai et~al.(2023)Bai, Bai, Yang, Wang, Tan, Wang, Lin, Zhou, and Zhou]{bai2023qwen}
Jinze Bai, Shuai Bai, Shusheng Yang, Shijie Wang, Sinan Tan, Peng Wang, Junyang Lin, Chang Zhou, and Jingren Zhou.
\newblock Qwen-vl: A versatile vision-language model for understanding, localization, text reading, and beyond.
\newblock \emph{arXiv preprint arXiv:2308.12966}, 2023.

\bibitem[Bhattacharya et~al.(2023)Bhattacharya, Singla, Krishnamurthy, Shah, and Chen]{bhattacharya2023video}
Aanisha Bhattacharya, Yaman~K Singla, Balaji Krishnamurthy, Rajiv~Ratn Shah, and Changyou Chen.
\newblock A video is worth 4096 tokens: Verbalize story videos to understand them in zero shot.
\newblock \emph{arXiv preprint arXiv:2305.09758}, 2023.

\bibitem[Brohan et~al.(2023)Brohan, Brown, Carbajal, Chebotar, Chen, Choromanski, Ding, Driess, Dubey, Finn, et~al.]{brohan2023rt}
Anthony Brohan, Noah Brown, Justice Carbajal, Yevgen Chebotar, Xi Chen, Krzysztof Choromanski, Tianli Ding, Danny Driess, Avinava Dubey, Chelsea Finn, et~al.
\newblock Rt-2: Vision-language-action models transfer web knowledge to robotic control.
\newblock \emph{arXiv preprint arXiv:2307.15818}, 2023.

\bibitem[Brown et~al.(2020)Brown, Mann, Ryder, Subbiah, Kaplan, Dhariwal, Neelakantan, Shyam, Sastry, Askell, et~al.]{gpt3}
Tom Brown, Benjamin Mann, Nick Ryder, Melanie Subbiah, Jared~D Kaplan, Prafulla Dhariwal, Arvind Neelakantan, Pranav Shyam, Girish Sastry, Amanda Askell, et~al.
\newblock Language models are few-shot learners.
\newblock \emph{Adv. Neural Inform. Process. Syst.}, pages 1877--1901, 2020.

\bibitem[Chen et~al.(2023)Chen, Zhu, Haydarov, Li, and Elhoseiny]{chen2023video}
Jun Chen, Deyao Zhu, Kilichbek Haydarov, Xiang Li, and Mohamed Elhoseiny.
\newblock Video chatcaptioner: Towards the enriched spatiotemporal descriptions.
\newblock \emph{arXiv preprint arXiv:2304.04227}, 2023.

\bibitem[Cheng et~al.(2023)Cheng, Wang, Lei, Crandall, Bansal, and Bertasius]{VindLU_CVPR2023}
Feng Cheng, Xizi Wang, Jie Lei, David Crandall, Mohit Bansal, and Gedas Bertasius.
\newblock Vindlu: A recipe for effective video-and-language pretraining.
\newblock In \emph{IEEE/CVF Conf. Comput. Vis. Pattern Recog.}, pages 10739--10750, 2023.

\bibitem[Choudhury et~al.(2023)Choudhury, Niinuma, Kitani, and Jeni]{choudhury2023zero}
Rohan Choudhury, Koichiro Niinuma, Kris~M Kitani, and L{\'a}szl{\'o}~A Jeni.
\newblock Zero-shot video question answering with procedural programs.
\newblock \emph{arXiv preprint arXiv:2312.00937}, 2023.

\bibitem[Chowdhery et~al.(2022)Chowdhery, Narang, Devlin, Bosma, Mishra, Roberts, Barham, Chung, Sutton, Gehrmann, et~al.]{chowdhery2022palm}
Aakanksha Chowdhery, Sharan Narang, Jacob Devlin, Maarten Bosma, Gaurav Mishra, Adam Roberts, Paul Barham, Hyung~Won Chung, Charles Sutton, Sebastian Gehrmann, et~al.
\newblock Palm: Scaling language modeling with pathways.
\newblock \emph{arXiv preprint arXiv:2204.02311}, 2022.

\bibitem[Chung and Yu(2023)]{chung2023long}
Jiwan Chung and Youngjae Yu.
\newblock Long story short: a summarize-then-search method for long video question answering.
\newblock In \emph{Brit. Mach. Vis. Conf.}, 2023.

\bibitem[Ding et~al.(2024)Ding, Zhang, Zhang, Xu, Shang, Xu, Yang, and Yang]{ding2024longrope}
Yiran Ding, Li~Lyna Zhang, Chengruidong Zhang, Yuanyuan Xu, Ning Shang, Jiahang Xu, Fan Yang, and Mao Yang.
\newblock Longrope: Extending llm context window beyond 2 million tokens.
\newblock \emph{arXiv preprint arXiv:2402.13753}, 2024.

\bibitem[Driess et~al.(2023)Driess, Xia, Sajjadi, Lynch, Chowdhery, Ichter, Wahid, Tompson, Vuong, Yu, et~al.]{driess2023palm}
Danny Driess, Fei Xia, Mehdi~SM Sajjadi, Corey Lynch, Aakanksha Chowdhery, Brian Ichter, Ayzaan Wahid, Jonathan Tompson, Quan Vuong, Tianhe Yu, et~al.
\newblock Palm-e: An embodied multimodal language model.
\newblock \emph{arXiv preprint arXiv:2303.03378}, 2023.

\bibitem[Fan et~al.(2024)Fan, Ma, Wu, Du, Li, Gao, and Li]{fan2024videoagent}
Yue Fan, Xiaojian Ma, Rujie Wu, Yuntao Du, Jiaqi Li, Zhi Gao, and Qing Li.
\newblock Videoagent: A memory-augmented multimodal agent for video understanding.
\newblock \emph{arXiv preprint arXiv:2403.11481}, 2024.

\bibitem[Fu et~al.(2024{\natexlab{a}})Fu, Dai, Luo, Li, Ren, Zhang, Wang, Zhou, Shen, Zhang, et~al.]{fu2024video}
Chaoyou Fu, Yuhan Dai, Yondong Luo, Lei Li, Shuhuai Ren, Renrui Zhang, Zihan Wang, Chenyu Zhou, Yunhang Shen, Mengdan Zhang, et~al.
\newblock Video-mme: The first-ever comprehensive evaluation benchmark of multi-modal llms in video analysis.
\newblock \emph{arXiv preprint arXiv:2405.21075}, 2024{\natexlab{a}}.

\bibitem[Fu et~al.(2024{\natexlab{b}})Fu, Lin, Long, Shen, Zhao, Zhang, Wang, Yin, Ma, Zheng, et~al.]{fu2024vita}
Chaoyou Fu, Haojia Lin, Zuwei Long, Yunhang Shen, Meng Zhao, Yifan Zhang, Xiong Wang, Di Yin, Long Ma, Xiawu Zheng, et~al.
\newblock Vita: Towards open-source interactive omni multimodal llm.
\newblock \emph{arXiv preprint arXiv:2408.05211}, 2024{\natexlab{b}}.

\bibitem[Gao et~al.(2023)Gao, Ji, Zhou, Lin, Chen, Fan, and Shou]{gao2023assistgpt}
Difei Gao, Lei Ji, Luowei Zhou, Kevin~Qinghong Lin, Joya Chen, Zihan Fan, and Mike~Zheng Shou.
\newblock Assistgpt: A general multi-modal assistant that can plan, execute, inspect, and learn.
\newblock \emph{arXiv preprint arXiv:2306.08640}, 2023.

\bibitem[Grauman et~al.(2022)Grauman, Westbury, Byrne, Chavis, Furnari, Girdhar, Hamburger, Jiang, Liu, Liu, et~al.]{grauman2022ego4d}
Kristen Grauman, Andrew Westbury, Eugene Byrne, Zachary Chavis, Antonino Furnari, Rohit Girdhar, Jackson Hamburger, Hao Jiang, Miao Liu, Xingyu Liu, et~al.
\newblock Ego4d: Around the world in 3,000 hours of egocentric video.
\newblock In \emph{IEEE/CVF Conf. Comput. Vis. Pattern Recog.}, pages 18995--19012, 2022.

\bibitem[Gu et~al.(2021)Gu, Goel, and R{\'e}]{gu2021efficiently}
Albert Gu, Karan Goel, and Christopher R{\'e}.
\newblock Efficiently modeling long sequences with structured state spaces.
\newblock \emph{arXiv preprint arXiv:2111.00396}, 2021.

\bibitem[Guo et~al.(2024)Guo, Zhu, Yang, Xie, Dong, Zhang, Chen, Bi, Wu, Li, et~al.]{guo2024deepseek}
Daya Guo, Qihao Zhu, Dejian Yang, Zhenda Xie, Kai Dong, Wentao Zhang, Guanting Chen, Xiao Bi, Yu Wu, YK Li, et~al.
\newblock Deepseek-coder: When the large language model meets programming--the rise of code intelligence.
\newblock \emph{arXiv preprint arXiv:2401.14196}, 2024.

\bibitem[Hong et~al.(2023)Hong, Wang, Lv, Xu, Yu, Ji, Wang, Wang, Dong, Ding, et~al.]{hong2023cogagent}
Wenyi Hong, Weihan Wang, Qingsong Lv, Jiazheng Xu, Wenmeng Yu, Junhui Ji, Yan Wang, Zihan Wang, Yuxiao Dong, Ming Ding, et~al.
\newblock Cogagent: A visual language model for gui agents.
\newblock \emph{arXiv preprint arXiv:2312.08914}, 2023.

\bibitem[Hussein et~al.(2019)Hussein, Gavves, and Smeulders]{hussein2019videograph}
Noureldien Hussein, Efstratios Gavves, and Arnold~WM Smeulders.
\newblock Videograph: Recognizing minutes-long human activities in videos.
\newblock \emph{arXiv preprint arXiv:1905.05143}, 2019.

\bibitem[Islam and Bertasius(2022)]{islam2022long}
Md~Mohaiminul Islam and Gedas Bertasius.
\newblock Long movie clip classification with state-space video models.
\newblock In \emph{Eur. Conf. Comput. Vis.}, pages 87--104, 2022.

\bibitem[Jeoung et~al.(2024)Jeoung, Huybrechts, Ganesh, Galstyan, and Bodapati]{jeoung2024adaptive}
Sullam Jeoung, Goeric Huybrechts, Bhavana Ganesh, Aram Galstyan, and Sravan Bodapati.
\newblock Adaptive video understanding agent: Enhancing efficiency with dynamic frame sampling and feedback-driven reasoning.
\newblock \emph{arXiv preprint arXiv:2410.20252}, 2024.

\bibitem[Jiang et~al.(2023)Jiang, Sablayrolles, Mensch, Bamford, Chaplot, Casas, Bressand, Lengyel, Lample, Saulnier, et~al.]{jiang2023mistral}
Albert~Q Jiang, Alexandre Sablayrolles, Arthur Mensch, Chris Bamford, Devendra~Singh Chaplot, Diego de~las Casas, Florian Bressand, Gianna Lengyel, Guillaume Lample, Lucile Saulnier, et~al.
\newblock Mistral 7b.
\newblock \emph{arXiv preprint arXiv:2310.06825}, 2023.

\bibitem[Kim et~al.(2024)Kim, Choi, Lee, and Rhee]{kim2024image}
Wonkyun Kim, Changin Choi, Wonseok Lee, and Wonjong Rhee.
\newblock An image grid can be worth a video: Zero-shot video question answering using a vlm.
\newblock \emph{arXiv preprint arXiv:2403.18406}, 2024.

\bibitem[Lei et~al.(2021)Lei, Li, Zhou, Gan, Berg, Bansal, and Liu]{lei2021less}
Jie Lei, Linjie Li, Luowei Zhou, Zhe Gan, Tamara~L Berg, Mohit Bansal, and Jingjing Liu.
\newblock Less is more: Clipbert for video-and-language learning via sparse sampling.
\newblock In \emph{IEEE/CVF Conf. Comput. Vis. Pattern Recog.}, pages 7331--7341, 2021.

\bibitem[Li et~al.(2024)Li, Zhang, Guo, Zhang, Li, Zhang, Zhang, Li, Liu, and Li]{li2024llava}
Bo Li, Yuanhan Zhang, Dong Guo, Renrui Zhang, Feng Li, Hao Zhang, Kaichen Zhang, Yanwei Li, Ziwei Liu, and Chunyuan Li.
\newblock Llava-onevision: Easy visual task transfer.
\newblock \emph{arXiv preprint arXiv:2408.03326}, 2024.

\bibitem[Li et~al.(2023{\natexlab{a}})Li, Li, Savarese, and Hoi]{li2023blip2}
Junnan Li, Dongxu Li, Silvio Savarese, and Steven Hoi.
\newblock {BLIP-2:} bootstrapping language-image pre-training with frozen image encoders and large language models.
\newblock In \emph{Proc. Int. Conf. Mach. Learn.}, 2023{\natexlab{a}}.

\bibitem[Li et~al.(2023{\natexlab{b}})Li, He, Wang, Li, Wang, Luo, Wang, Wang, and Qiao]{li2023videochat}
KunChang Li, Yinan He, Yi Wang, Yizhuo Li, Wenhai Wang, Ping Luo, Yali Wang, Limin Wang, and Yu Qiao.
\newblock Videochat: Chat-centric video understanding.
\newblock \emph{arXiv preprint arXiv:2305.06355}, 2023{\natexlab{b}}.

\bibitem[Lin et~al.(2023{\natexlab{a}})Lin, Zhu, Ye, Ning, Jin, and Yuan]{videoLlava}
Bin Lin, Bin Zhu, Yang Ye, Munan Ning, Peng Jin, and Li Yuan.
\newblock Video-llava: Learning united visual representation by alignment before projection.
\newblock \emph{arXiv preprint arXiv:2311.10122}, 2023{\natexlab{a}}.

\bibitem[Lin et~al.(2023{\natexlab{b}})Lin, Ahmed, Li, Lin, Azarnasab, Yang, Wang, Liang, Liu, Lu, et~al.]{lin2023mm}
Kevin Lin, Faisal Ahmed, Linjie Li, Chung-Ching Lin, Ehsan Azarnasab, Zhengyuan Yang, Jianfeng Wang, Lin Liang, Zicheng Liu, Yumao Lu, et~al.
\newblock Mm-vid: Advancing video understanding with gpt-4v (ision).
\newblock \emph{arXiv preprint arXiv:2310.19773}, 2023{\natexlab{b}}.

\bibitem[Liu et~al.(2024)Liu, Li, Li, Li, Zhang, Shen, and Lee]{liu2024llavanext}
Haotian Liu, Chunyuan Li, Yuheng Li, Bo Li, Yuanhan Zhang, Sheng Shen, and Yong~Jae Lee.
\newblock Llava-next: Improved reasoning, ocr, and world knowledge.
\newblock \url{https://llava-vl.github.io/blog/2024-01-30-llava-next/}, 2024.

\bibitem[Ma et~al.(2024)Ma, Li, Sun, Cai, Long, and Ma]{ma2024gerea}
Ziyu Ma, Shutao Li, Bin Sun, Jianfei Cai, Zuxiang Long, and Fuyan Ma.
\newblock Gerea: Question-aware prompt captions for knowledge-based visual question answering.
\newblock \emph{arXiv preprint arXiv:2402.02503}, 2024.

\bibitem[Maaz et~al.(2023)Maaz, Rasheed, Khan, and Khan]{maaz2023video}
Muhammad Maaz, Hanoona Rasheed, Salman Khan, and Fahad~Shahbaz Khan.
\newblock Video-chatgpt: Towards detailed video understanding via large vision and language models.
\newblock \emph{arXiv preprint arXiv:2306.05424}, 2023.

\bibitem[Mangalam et~al.(2023)Mangalam, Akshulakov, and Malik]{mangalam2023egoschema}
Karttikeya Mangalam, Raiymbek Akshulakov, and Jitendra Malik.
\newblock Egoschema: A diagnostic benchmark for very long-form video language understanding.
\newblock \emph{arXiv preprint arXiv:2308.09126}, 2023.

\bibitem[Min et~al.(2024)Min, Buch, Nagrani, Cho, and Schmid]{min2024morevqa}
Juhong Min, Shyamal Buch, Arsha Nagrani, Minsu Cho, and Cordelia Schmid.
\newblock Morevqa: Exploring modular reasoning models for video question answering.
\newblock In \emph{IEEE/CVF Conf. Comput. Vis. Pattern Recog.}, pages 13235--13245, 2024.

\bibitem[Muhammad~Maaz and Khan(2023)]{videoChatGpt}
Salman~Khan Muhammad~Maaz, Hanoona~Rasheed and Fahad Khan.
\newblock Video-chatgpt: Towards detailed video understanding via large vision and language models.
\newblock \emph{ArXiv 2306.05424}, 2023.

\bibitem[Nguyen et~al.(2022)Nguyen, Goel, Gu, Downs, Shah, Dao, Baccus, and R{\'e}]{nguyen2022s4nd}
Eric Nguyen, Karan Goel, Albert Gu, Gordon Downs, Preey Shah, Tri Dao, Stephen Baccus, and Christopher R{\'e}.
\newblock S4nd: Modeling images and videos as multidimensional signals with state spaces.
\newblock \emph{Adv. Neural Inform. Process. Syst.}, pages 2846--2861, 2022.

\bibitem[OpenAI(2021)]{gpt3.5}
OpenAI.
\newblock {GPT3.5}.
\newblock \url{https://platform.openai.com/docs/models/gpt-3-5}, 2021.

\bibitem[OpenAI(2022)]{chatgpt}
OpenAI.
\newblock Introducing chatgpt.
\newblock \url{https://openai.com/blog/chatgpt}, 2022.

\bibitem[OpenAI(2023)]{openai2023gpt4v}
OpenAI.
\newblock {GPT-4V(ision) System Card}.
\newblock \url{https://openai.com/index/gpt-4v-system-card/}, 2023.

\bibitem[OpenAI(2024{\natexlab{a}})]{openai2024}
OpenAI.
\newblock New embedding models and api updates.
\newblock \url{https://openai.com/index/new-embedding-models-and-api-updates/}, 2024{\natexlab{a}}.

\bibitem[OpenAI(2024{\natexlab{b}})]{openai2024gpt4o}
OpenAI.
\newblock {GPT-4o Mini: Advancing Cost-Efficient Intelligence}.
\newblock \url{https://openai.com/index/gpt-4o-mini-advancing-cost-efficient-intelligence/}, 2024{\natexlab{b}}.

\bibitem[Ouyang et~al.(2022)Ouyang, Wu, Jiang, Almeida, Wainwright, Mishkin, Zhang, Agarwal, Slama, Ray, et~al.]{instructGPT}
Long Ouyang, Jeffrey Wu, Xu Jiang, Diogo Almeida, Carroll Wainwright, Pamela Mishkin, Chong Zhang, Sandhini Agarwal, Katarina Slama, Alex Ray, et~al.
\newblock Training language models to follow instructions with human feedback.
\newblock \emph{Adv. Neural Inform. Process. Syst.}, pages 27730--27744, 2022.

\bibitem[Perot et~al.(2023)Perot, Kang, Luisier, Su, Sun, Boppana, Wang, Wang, Mu, Zhang, et~al.]{perot2023lmdx}
Vincent Perot, Kai Kang, Florian Luisier, Guolong Su, Xiaoyu Sun, Ramya~Sree Boppana, Zilong Wang, Zifeng Wang, Jiaqi Mu, Hao Zhang, et~al.
\newblock Lmdx: Language model-based document information extraction and localization.
\newblock \emph{arXiv preprint arXiv:2309.10952}, 2023.

\bibitem[Ranasinghe et~al.(2024)Ranasinghe, Li, Kahatapitiya, and Ryoo]{ranasinghe2024understanding}
Kanchana Ranasinghe, Xiang Li, Kumara Kahatapitiya, and Michael~S Ryoo.
\newblock Understanding long videos in one multimodal language model pass.
\newblock \emph{arXiv preprint arXiv:2403.16998}, 2024.

\bibitem[Scao et~al.(2022)Scao, Fan, Akiki, Pavlick, Ili{\'c}, Hesslow, Castagn{\'e}, Luccioni, Yvon, Gall{\'e}, et~al.]{bloom}
Teven~Le Scao, Angela Fan, Christopher Akiki, Ellie Pavlick, Suzana Ili{\'c}, Daniel Hesslow, Roman Castagn{\'e}, Alexandra~Sasha Luccioni, Fran{\c{c}}ois Yvon, Matthias Gall{\'e}, et~al.
\newblock Bloom: A 176b-parameter open-access multilingual language model.
\newblock \emph{arXiv preprint arXiv:2211.05100}, 2022.

\bibitem[Song et~al.(2023)Song, Chai, Wang, Zhang, Zhou, Wu, Guo, Ye, Lu, Hwang, et~al.]{song2023moviechat}
Enxin Song, Wenhao Chai, Guanhong Wang, Yucheng Zhang, Haoyang Zhou, Feiyang Wu, Xun Guo, Tian Ye, Yan Lu, Jenq-Neng Hwang, et~al.
\newblock Moviechat: From dense token to sparse memory for long video understanding.
\newblock \emph{arXiv preprint arXiv:2307.16449}, 2023.

\bibitem[Song et~al.(2024)Song, Chai, Ye, Hwang, Li, and Wang]{song2024moviechat+}
Enxin Song, Wenhao Chai, Tian Ye, Jenq-Neng Hwang, Xi Li, and Gaoang Wang.
\newblock Moviechat+: Question-aware sparse memory for long video question answering.
\newblock \emph{arXiv preprint arXiv:2404.17176}, 2024.

\bibitem[Sun et~al.(2024)Sun, Wang, Yu, Cui, Zhang, Zhang, and Wang]{sun2024eva}
Quan Sun, Jinsheng Wang, Qiying Yu, Yufeng Cui, Fan Zhang, Xiaosong Zhang, and Xinlong Wang.
\newblock Eva-clip-18b: Scaling clip to 18 billion parameters.
\newblock \emph{arXiv preprint arXiv:2402.04252}, 2024.

\bibitem[Sun et~al.(2022)Sun, Xue, Song, Liu, Yang, and Fu]{sun2022long}
Yuchong Sun, Hongwei Xue, Ruihua Song, Bei Liu, Huan Yang, and Jianlong Fu.
\newblock Long-form video-language pre-training with multimodal temporal contrastive learning.
\newblock \emph{Adv. Neural Inform. Process. Syst.}, 35:\penalty0 38032--38045, 2022.

\bibitem[Sur\'is et~al.(2023)Sur\'is, Menon, and Vondrick]{suris2023vipergpt}
D\'idac Sur\'is, Sachit Menon, and Carl Vondrick.
\newblock Vipergpt: Visual inference via python execution for reasoning.
\newblock \emph{Int. Conf. Comput. Vis.}, pages 11888--11898, 2023.

\bibitem[Team et~al.(2023)Team, Anil, Borgeaud, Wu, Alayrac, Yu, Soricut, Schalkwyk, Dai, Hauth, et~al.]{team2023gemini}
Gemini Team, Rohan Anil, Sebastian Borgeaud, Yonghui Wu, Jean-Baptiste Alayrac, Jiahui Yu, Radu Soricut, Johan Schalkwyk, Andrew~M Dai, Anja Hauth, et~al.
\newblock Gemini: a family of highly capable multimodal models.
\newblock \emph{arXiv preprint arXiv:2312.11805}, 2023.

\bibitem[Team et~al.(2024)Team, Georgiev, Lei, Burnell, Bai, Gulati, Tanzer, Vincent, Pan, Wang, et~al.]{team2024gemini}
Gemini Team, Petko Georgiev, Ving~Ian Lei, Ryan Burnell, Libin Bai, Anmol Gulati, Garrett Tanzer, Damien Vincent, Zhufeng Pan, Shibo Wang, et~al.
\newblock Gemini 1.5: Unlocking multimodal understanding across millions of tokens of context.
\newblock \emph{arXiv preprint arXiv:2403.05530}, 2024.

\bibitem[Touvron et~al.(2023)Touvron, Lavril, Izacard, Martinet, Lachaux, Lacroix, Rozi{\`e}re, Goyal, Hambro, Azhar, et~al.]{llama}
Hugo Touvron, Thibaut Lavril, Gautier Izacard, Xavier Martinet, Marie-Anne Lachaux, Timoth{\'e}e Lacroix, Baptiste Rozi{\`e}re, Naman Goyal, Eric Hambro, Faisal Azhar, et~al.
\newblock Llama: Open and efficient foundation language models.
\newblock \emph{arXiv preprint arXiv:2302.13971}, 2023.

\bibitem[Wang et~al.(2023{\natexlab{a}})Wang, Chen, Luo, Dai, Yuan, Wu, and Jiang]{wang2023chatvideo}
Junke Wang, Dongdong Chen, Chong Luo, Xiyang Dai, Lu Yuan, Zuxuan Wu, and Yu-Gang Jiang.
\newblock Chatvideo: A tracklet-centric multimodal and versatile video understanding system.
\newblock \emph{arXiv preprint arXiv:2304.14407}, 2023{\natexlab{a}}.

\bibitem[Wang et~al.(2023{\natexlab{b}})Wang, Zhu, Wang, Yu, Liu, Omar, and Hamid]{wang2023selective}
Jue Wang, Wentao Zhu, Pichao Wang, Xiang Yu, Linda Liu, Mohamed Omar, and Raffay Hamid.
\newblock Selective structured state-spaces for long-form video understanding.
\newblock In \emph{IEEE/CVF Conf. Comput. Vis. Pattern Recog.}, pages 6387--6397, 2023{\natexlab{b}}.

\bibitem[Wang et~al.(2024{\natexlab{a}})Wang, Yuan, and Zhang]{wang2024tarsier}
Jiawei Wang, Liping Yuan, and Yuchen Zhang.
\newblock Tarsier: Recipes for training and evaluating large video description models.
\newblock \emph{arXiv preprint arXiv:2407.00634}, 2024{\natexlab{a}}.

\bibitem[Wang et~al.(2023{\natexlab{c}})Wang, Zhao, Do, Agarwal, Lee, and Sun]{wang2023vamos}
Shijie Wang, Qi Zhao, Minh~Quan Do, Nakul Agarwal, Kwonjoon Lee, and Chen Sun.
\newblock Vamos: Versatile action models for video understanding.
\newblock \emph{arXiv preprint arXiv:2311.13627}, 2023{\natexlab{c}}.

\bibitem[Wang et~al.(2024{\natexlab{b}})Wang, Zhang, Zohar, and Yeung-Levy]{wang2024videoagent}
Xiaohan Wang, Yuhui Zhang, Orr Zohar, and Serena Yeung-Levy.
\newblock Videoagent: Long-form video understanding with large language model as agent.
\newblock \emph{arXiv preprint arXiv:2403.10517}, 2024{\natexlab{b}}.

\bibitem[Wang et~al.(2021)Wang, Bertasius, Oh, Gupta, Hoai, and Torresani]{wang2021supervoxel}
Yang Wang, Gedas Bertasius, Tae-Hyun Oh, Abhinav Gupta, Minh Hoai, and Lorenzo Torresani.
\newblock Supervoxel attention graphs for long-range video modeling.
\newblock In \emph{Proc. IEEE/CVF Winter Conf. Appl. Comput. Vis.}, pages 155--166, 2021.

\bibitem[Wang et~al.(2023{\natexlab{d}})Wang, He, Li, Li, Yu, Ma, Li, Chen, Chen, Wang, et~al.]{wang2023internvid}
Yi Wang, Yinan He, Yizhuo Li, Kunchang Li, Jiashuo Yu, Xin Ma, Xinhao Li, Guo Chen, Xinyuan Chen, Yaohui Wang, et~al.
\newblock Internvid: A large-scale video-text dataset for multimodal understanding and generation.
\newblock \emph{arXiv preprint arXiv:2307.06942}, 2023{\natexlab{d}}.

\bibitem[Wang et~al.(2022)Wang, Li, Xu, Zhou, Lei, Lin, Wang, Yang, Zhu, Hoiem, et~al.]{wang2022language}
Zhenhailong Wang, Manling Li, Ruochen Xu, Luowei Zhou, Jie Lei, Xudong Lin, Shuohang Wang, Ziyi Yang, Chenguang Zhu, Derek Hoiem, et~al.
\newblock Language models with image descriptors are strong few-shot video-language learners.
\newblock \emph{Adv. Neural Inform. Process. Syst.}, pages 8483--8497, 2022.

\bibitem[Wei et~al.(2022)Wei, Wang, Schuurmans, Bosma, Xia, Chi, Le, Zhou, et~al.]{wei2022chain}
Jason Wei, Xuezhi Wang, Dale Schuurmans, Maarten Bosma, Fei Xia, Ed Chi, Quoc~V Le, Denny Zhou, et~al.
\newblock Chain-of-thought prompting elicits reasoning in large language models.
\newblock \emph{Adv. Neural Inform. Process. Syst.}, pages 24824--24837, 2022.

\bibitem[Wu et~al.(2022)Wu, Li, Mangalam, Fan, Xiong, Malik, and Feichtenhofer]{wu2022memvit}
Chao-Yuan Wu, Yanghao Li, Karttikeya Mangalam, Haoqi Fan, Bo Xiong, Jitendra Malik, and Christoph Feichtenhofer.
\newblock Memvit: Memory-augmented multiscale vision transformer for efficient long-term video recognition.
\newblock In \emph{IEEE/CVF Conf. Comput. Vis. Pattern Recog.}, pages 13587--13597, 2022.

\bibitem[Xu et~al.(2024)Xu, Zhao, Zhou, Lin, Ng, and Feng]{xu2024pllava}
Lin Xu, Yilin Zhao, Daquan Zhou, Zhijie Lin, See~Kiong Ng, and Jiashi Feng.
\newblock Pllava: Parameter-free llava extension from images to videos for video dense captioning.
\newblock \emph{arXiv preprint arXiv:2404.16994}, 2024.

\bibitem[Yang et~al.(2021)Yang, Miech, Sivic, Laptev, and Schmid]{yang2021just}
Antoine Yang, Antoine Miech, Josef Sivic, Ivan Laptev, and Cordelia Schmid.
\newblock Just ask: Learning to answer questions from millions of narrated videos.
\newblock In \emph{Int. Conf. Comput. Vis.}, pages 1686--1697, 2021.

\bibitem[Yang et~al.(2022)Yang, Miech, Sivic, Laptev, and Schmid]{yang2022zero}
Antoine Yang, Antoine Miech, Josef Sivic, Ivan Laptev, and Cordelia Schmid.
\newblock Zero-shot video question answering via frozen bidirectional language models.
\newblock \emph{Adv. Neural Inform. Process. Syst.}, 35:\penalty0 124--141, 2022.

\bibitem[Yang et~al.(2023)Yang, Chu, Feiszli, Goyal, Torresani, and Tran]{yang2023relational}
Xitong Yang, Fu-Jen Chu, Matt Feiszli, Raghav Goyal, Lorenzo Torresani, and Du Tran.
\newblock Relational space-time query in long-form videos.
\newblock In \emph{IEEE/CVF Conf. Comput. Vis. Pattern Recog.}, pages 6398--6408, 2023.

\bibitem[Yang et~al.(2024)Yang, Chen, Li, Wang, and Yang]{yang2024doraemongpt}
Zongxin Yang, Guikun Chen, Xiaodi Li, Wenguan Wang, and Yi Yang.
\newblock Doraemongpt: Toward understanding dynamic scenes with large language models.
\newblock \emph{arXiv preprint arXiv:2401.08392}, 2024.

\bibitem[Zeng et~al.(2022)Zeng, Attarian, Ichter, Choromanski, Wong, Welker, Tombari, Purohit, Ryoo, Sindhwani, Lee, Vanhoucke, and Florence]{zeng2022socraticmodels}
Andy Zeng, Maria Attarian, Brian Ichter, Krzysztof Choromanski, Adrian Wong, Stefan Welker, Federico Tombari, Aveek Purohit, Michael Ryoo, Vikas Sindhwani, Johnny Lee, Vincent Vanhoucke, and Pete Florence.
\newblock Socratic models: Composing zero-shot multimodal reasoning with language.
\newblock \emph{arXiv preprint arXiv:2204.00598}, 2022.

\bibitem[Zhang et~al.(2023{\natexlab{a}})Zhang, Lu, Islam, Wang, Yu, Bansal, and Bertasius]{LLoVi}
Ce Zhang, Taixi Lu, Md~Mohaiminul Islam, Ziyang Wang, Shoubin Yu, Mohit Bansal, and Gedas Bertasius.
\newblock A simple llm framework for long-range video question-answering.
\newblock \emph{arXiv preprint arXiv:2312.17235}, 2023{\natexlab{a}}.

\bibitem[Zhang et~al.(2023{\natexlab{b}})Zhang, Li, and Bing]{video-llama}
Hang Zhang, Xin Li, and Lidong Bing.
\newblock Video-llama: An instruction-tuned audio-visual language model for video understanding.
\newblock \emph{arXiv preprint arXiv:2306.02858}, 2023{\natexlab{b}}.

\bibitem[Zhang et~al.(2024{\natexlab{a}})Zhang, Zhang, Li, Zeng, Yang, Zhang, Wang, Tan, Li, and Liu]{zhang2024long}
Peiyuan Zhang, Kaichen Zhang, Bo Li, Guangtao Zeng, Jingkang Yang, Yuanhan Zhang, Ziyue Wang, Haoran Tan, Chunyuan Li, and Ziwei Liu.
\newblock Long context transfer from language to vision.
\newblock \emph{arXiv preprint arXiv:2406.16852}, 2024{\natexlab{a}}.

\bibitem[Zhang et~al.(2024{\natexlab{b}})Zhang, Li, Liu, Lee, Gui, Fu, Feng, Liu, and Li]{zhang2024llavanextvideo}
Yuanhan Zhang, Bo Li, Haotian Liu, Yong~Jae Lee, Liangke Gui, Di Fu, Jiashi Feng, Ziwei Liu, and Chunyuan Li.
\newblock Llava-next: A strong zero-shot video understanding model.
\newblock \url{https://llava-vl.github.io/blog/2024-04-30-llava-next-video/}, 2024{\natexlab{b}}.

\bibitem[Zhao et~al.(2023)Zhao, Misra, Kr{\"a}henb{\"u}hl, and Girdhar]{zhao2023learning}
Yue Zhao, Ishan Misra, Philipp Kr{\"a}henb{\"u}hl, and Rohit Girdhar.
\newblock Learning video representations from large language models.
\newblock In \emph{IEEE/CVF Conf. Comput. Vis. Pattern Recog.}, pages 6586--6597, 2023.

\end{thebibliography}
